\documentclass[10pt,twocolumn,letterpaper]{article}

\usepackage[pagenumbers]{cvpr} % To force page numbers, e.g. for an arXiv version
\usepackage[accsupp]{axessibility}
\usepackage{graphicx}
\usepackage{amsmath}
\usepackage{amssymb}
\usepackage{booktabs}
\usepackage{subcaption}

\usepackage{color}
\usepackage{colortbl}
\usepackage{pifont}
\usepackage{multirow}

\usepackage{lipsum}

\newcommand\blfootnote[1]{%
  \begingroup
  \renewcommand\thefootnote{}\footnote{#1}%
  \addtocounter{footnote}{-1}%
  \endgroup
}

\definecolor{turquoise}{cmyk}{0.65,0,0.1,0.3}
\definecolor{purple}{rgb}{0.65,0,0.65}
\definecolor{dark_green}{rgb}{0, 0.5, 0}
\definecolor{orange}{rgb}{0.8, 0.6, 0.2}
\definecolor{red}{rgb}{0.8, 0.2, 0.2}
\definecolor{darkred}{rgb}{0.6, 0.1, 0.05}
\definecolor{blueish}{rgb}{0.0, 0.3, .6}
\definecolor{light_gray}{rgb}{0.7, 0.7, .7}
\definecolor{pink}{rgb}{1, 0, 1}
\definecolor{greyblue}{rgb}{0.25, 0.25, 1}

\renewcommand{\paragraph}[1]{\vspace{1em}\noindent\textbf{#1}.}

\newcommand{\cmark}{\ding{51}}%
\newcommand{\xmark}{\ding{55}}%

\newcommand{\unaryminus}{\scalebox{0.75}[1.0]{\( - \)}}

\usepackage[pagebackref,breaklinks,colorlinks]{hyperref}
 \makeatletter
 \def\hlinewd#1{%
      \noalign{\ifnum0=`}\fi\hrule \@height #1 \futurelet
      \reserved@a\@xhline}

\usepackage[capitalize]{cleveref}
\crefname{section}{Sec.}{Secs.}
\Crefname{section}{Section}{Sections}
\Crefname{table}{Table}{Tables}
\crefname{table}{Tab.}{Tabs.}

\definecolor{light_gray}{gray}{0.9}
\definecolor{light-green}{rgb}{0.82, 0.94, 0.75}

\def\cvprPaperID{11772} % *** Enter the CVPR Paper ID here
\def\confName{CVPR}
\def\confYear{2022}

\begin{document}

%%%%%%%%% TITLE - PLEASE UPDATE
\title{Probabilistic Representations for Video Contrastive Learning}

% \author{First Author\\
% Institution1\\
% Institution1 address\\
% {\tt\small firstauthor@i1.org}
% % For a paper whose authors are all at the same institution,
% % omit the following lines up until the closing ``}''.
% % Additional authors and addresses can be added with ``\and'',
% % just like the second author.
% % To save space, use either the email address or home page, not both
% \and
% Second Author\\
% Institution2\\
% First line of institution2 address\\
% {\tt\small secondauthor@i2.org}
% }
\author{Jungin Park$^1$ \quad\quad\quad Jiyoung Lee$^2$ \quad\quad\quad Ig-Jae Kim$^3$ \quad\quad\quad Kwanghoon Sohn$^1$\thanks{Corresponding author.} \\
$^1$Yonsei University \quad\quad $^2$NAVER AI Lab \quad\quad
$^3$Korea Institute of Science and Technology (KIST)\\
{\tt\small $\lbrace$newrun, khsohn$\rbrace$@yonsei.ac.kr} \quad\quad\quad
\tt\small lee.j@navercorp.com}

\maketitle

\begin{abstract}
    \blfootnote{This research was supported by R\&D program for Advanced Integrated-intelligence for Identification (AIID) through the National Research Foundation of KOREA(NRF) funded by Ministry of Science and ICT (NRF-2018M3E3A1057289) and the Yonsei University Research Fund of 2021 (2021-22-0001).}
    This paper presents \textbf{Pro}babilistic \textbf{Vi}deo \textbf{Co}ntrastive Learning, a self-supervised representation learning method that bridges contrastive learning with probabilistic representation.
    We hypothesize that the clips composing the video have different distributions in short-term duration, but can represent the complicated and sophisticated video distribution through combination in a common embedding space.
    Thus, the proposed method represents video clips as normal distributions and combines them into a Mixture of Gaussians to model the whole video distribution.
    By sampling embeddings from the whole video distribution, we can circumvent the careful sampling strategy or transformations to generate augmented views of the clips, unlike previous deterministic methods that have mainly focused on such sample generation strategies for contrastive learning.
    We further propose a stochastic contrastive loss to learn proper video distributions and handle the inherent uncertainty from the nature of the raw video.
    Experimental results verify that our probabilistic embedding stands as a state-of-the-art video representation learning for action recognition and video retrieval on the most popular benchmarks, including UCF101 and HMDB51.
    
\end{abstract}

\vspace{-13pt}
\section{Introduction}
\label{sec:intro}\vspace{-5pt}
    
    Video is the vitality of the Internet, which means that understanding video content is essential for the most modern artificial intelligence (AI) agents.
    Alongside this, learning enriched spatiotemporal representations from \textit{unlabeled videos} (\ie, self-supervised or unsupervised video representation learning)~\cite{video-moco, cvrl, mfo} has become a crucial research topic for the computer vision community.
    The interest in this topic is to learn deep features representing general visual contents, which has proven essential to improving performance on downstream tasks such as action recognition~\cite{r3d, s3d, i3d}, action detection~\cite{action-detection1, action-detection2}, video retrieval~\cite{video-retrieval1, video-retrieval2}, and even multi-modal event recognition~\cite{multimodal-event, multimodal-event2}.
    However, self-supervised video representation learning has still remained challenging due to the inherent difficulty caused by the nature of the videos in comparison to static images.
    
    \begin{figure}[t]
        \centering
            \begin{subfigure}{1.0\linewidth}
           {\includegraphics[width=1.0\linewidth]{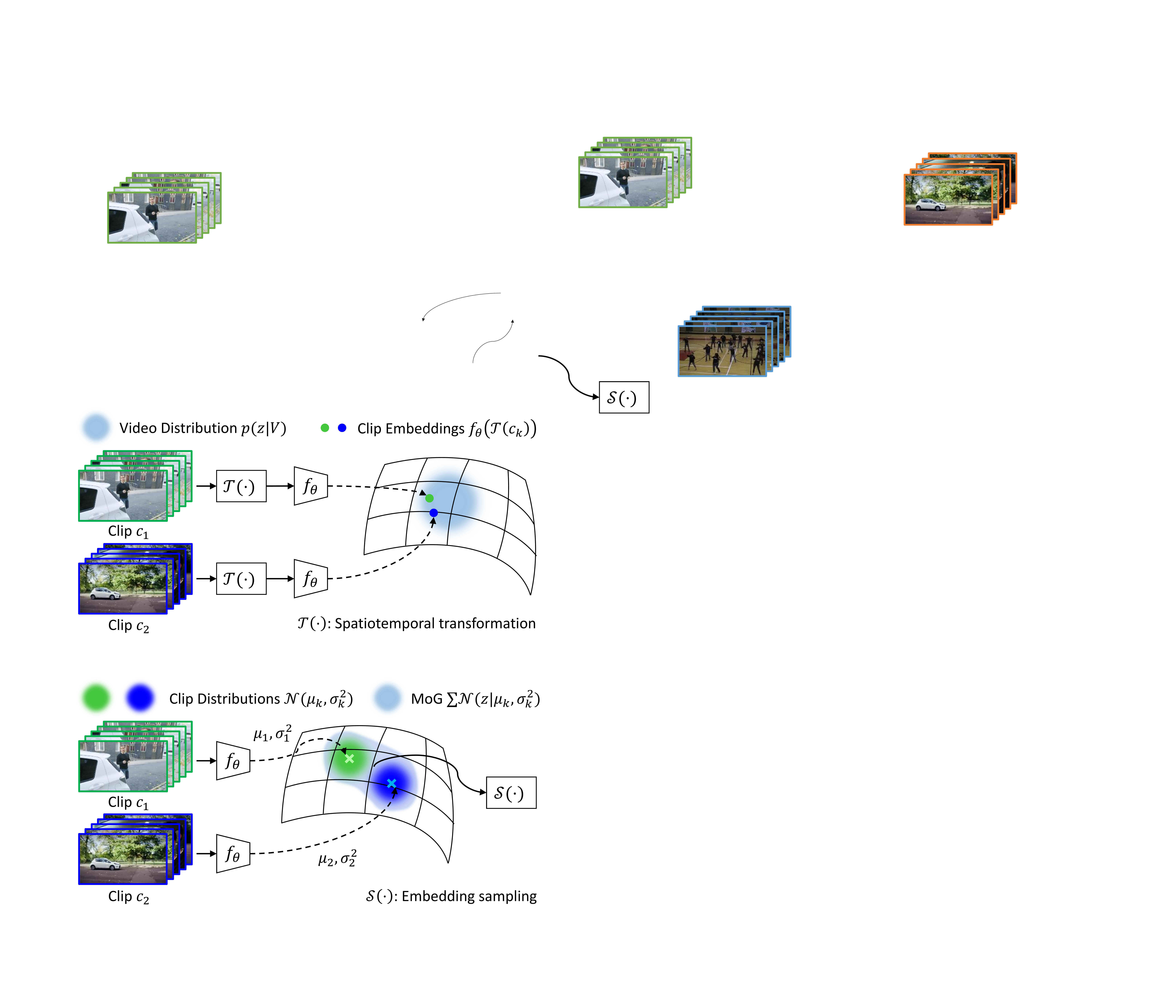}}\hfill
           \caption{Deterministic embedding}\label{fig:1a}
           \end{subfigure}
           \\ 
           \begin{subfigure}{1.0\linewidth}
           {\includegraphics[width=1.0\linewidth]{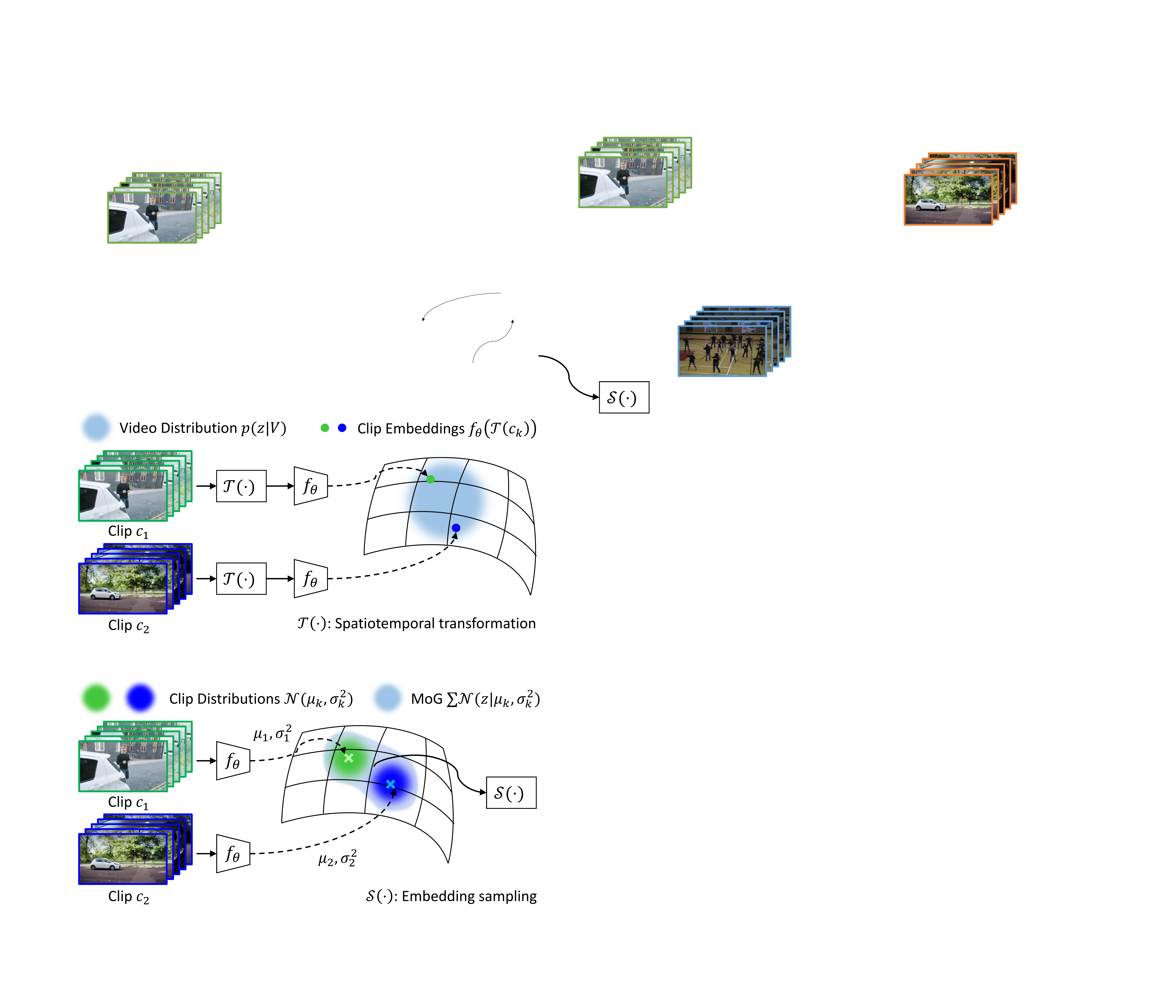}}\hfill 
           \caption{Probabilistic embedding (\textbf{Ours})}
           \end{subfigure}\label{fig:1b}\\
        \vspace{-5pt}
        \caption{
        Contrary to (a) the deterministic point embedding methods, which estimate the subset of the video distribution, (b) the proposed method estimates the whole video distribution through the mixture of probabilistic embeddings.}
    \label{fig:1}\vspace{-15pt}
    \end{figure}

    Recent breakthroughs in self-supervised video representation learning have been developed with two different branches: (1) leveraging pretext tasks related to the coherence of videos and (2) using contrastive learning~\cite{contrastive} for instance discrimination.
    Specifically, video coherence is empirically modeled through sub-properties of video contents associated with temporal ordering~\cite{sorting-sequences, odd-one-out, shuffle, vcop, rtt}, optical flow~\cite{flow-prediction}, spatiotemporal statistics~\cite{vcp, mas, stcp}, and playback rate~\cite{prp, pp}.
    Even though they have shown that spatiotemporal representations can be learned from unlabeled videos, the learned representations inevitably contain task-specific information.
    
    In contrast, instance discrimination methods~\cite{cvrl, video-moco, feichtenhofer21, cate, mcl, mcn} have attempted to learn video representations by incorporating contrastive learning~\cite{contrastive}, which aims to discriminate different instances without using sub-properties of data~\cite{contrastive-tpami, contrastive-cvpr}.
    Concretely, to learn spatiotemporal representations from videos, existing works treat each video as an ``instance" and embed video clips to deterministic points on the embedding space, as shown in \figref{fig:1a}.
    Based on contrastive objectives~\cite{contrastive-loss, ranking-loss, margin-loss, triplet-loss}, positive point pairs are pulled together and negative point pairs are pushed away.
    The positive pairs are composed of clips from the same video~\cite{cvrl} or different views (augmented versions) of the same clip~\cite{video-moco, feichtenhofer21}, and the negative pairs are composed of clips from different videos.
    To make such training pairs, several works have introduced carefully designed spatial and temporal transformations in the form of data augmentation, including a temporal mask~\cite{video-moco} and temporally consistent spatial augmentation~\cite{cvrl}.

    However, deterministic representations for video contrastive learning have critical limitations in three respects:
    First, representing the complicated and sophisticated video distribution as a set of deterministic points is insufficient to learn discriminative video representations.
    Unlike static images, videos are a collection of noisy temporal dynamics and contain a lot of redundant information, that makes the uncertainty of data high~\cite{active-contrastive}.
    Therefore, an alternative to deterministic representations is required to describe overall video distribution.
    Second, improper sampling and transformation techniques to generate different views can cause performance fluctuation according to downstream tasks~\cite{what-makes}.
    Moreover, improper temporal transformations (\eg shuffling) that can harm the video contents weaken the effectiveness of contrastive learning~\cite{taco}.
    Third, they often neglect common components that are likely to contain valid correspondences between semantically adjacent instances (\eg same category, but different videos), leading to limited discrimination performance of learned representations, as demonstrated in~\cite{co-crl}.

    To overcome these limitations while maximizing the advantages of contrastive learning, we propose probabilistic representations for video contrastive learning, named \textbf{ProViCo}, in which video clips are represented as random variables in a stochastic embedding space.
    As shown in \figref{fig:1b}, clips sampled from a video are represented as distinct normal distributions and the distribution of the whole video is approximated by a Mixture of Gaussians (MoG) of clip distributions.
    We construct the positive and negative pairs based on the probabilistic distance between embeddings sampled from each video distribution.
    Moreover, we propose an uncertainty-based stochastic contrastive loss that incorporates uncertainty (\ie, the inherent noise of videos) into the soft contrastive loss~\cite{hib}.
    By leveraging uncertainty, we can reduce the effect of noisy samples or improper training pairs on self-supervised representation learning, and can make useful applications such as estimation of difficulty or chance of failure on test data.
    
    To sum up, our contributions are as follows:
    (1) We propose a novel ProViCo to effectively represent the probabilistic video embedding space.
    To the best of our knowledge, this is the first attempt to leverage probabilistic embeddings for self-supervised video representation learning.
    (2) We introduce the probabilistic distance-based positive mining to exploit semantic relations between videos and present the stochastic contrastive loss to weaken the adverse impact of unreliable instances.
    (3) We demonstrate the effectiveness of the proposed probabilistic approach through the uncertainty estimation and extensive experiments on downstream tasks, including action recognition and video retrieval.
    
    \vspace{-10pt}
\section{Related Work}
\vspace{-15pt}
\paragraph{Self-supervised video representation learning}
    Early works on self-supervised video representation learning have been studied with pretext tasks to exploit the spatiotemporal cues, including prediction of motion and appearance~\cite{mas}, spatiotemporal transformations~\cite{shuffle, stcp, rtt, vcop}, frames~\cite{vcp}, and playback rate~\cite{prp, pp}.
    Recently, contrastive learning methods with instance discrimination tasks have been proposed for video representations~\cite{feichtenhofer21, cvrl, video-moco}.
    The popular self-supervised visual representation learning frameworks~\cite{moco, simclr, byol, swav} have been transformed to empower the temporal robustness of the encoder for video representations, improving momentum contrastive learning~\cite{feichtenhofer21, video-moco}.
    Further, motion estimation~\cite{mcl}, temporal relations~\cite{corp}, multi-level feature optimization~\cite{mfo}, and meta-learning framework~\cite{mcn} have been incorporated into contrastive learning.
    While several works have employed additional signals (\eg audio~\cite{audio-video1, audio-video2, audio-video3, active-contrastive, multisensory, stica} and optical flow~\cite{flow-prediction, co-crl, seco}) to improve the performance, we focus on RGB-only self-supervised video representation learning without growth of training costs following~\cite{video-moco, feichtenhofer21, mfo}.
    
\vspace{-10pt}
\paragraph{Probabilistic representation}
    Learning representations in a stochastic embedding space has first been proposed for word embeddings~\cite{word-embedding1}.
    Thanks to the high robustness of handling the inherent hierarchies of the language of probabilistic embeddings, they have been extensively explored in natural language processing~\cite{word-embedding2, word-embedding3}.
    The probabilistic embeddings for vision tasks have been introduced for face recognition~\cite{uncertainty-face2, uncertainty-face3}, speaker diarization~\cite{speaker-diarization}, human pose estimation~\cite{prob-humanpose}, and prototype embeddings for few-shot recognition~\cite{prob-prototype}.
    In recent years, hedged instance embedding (HIB)~\cite{hib} has been proposed to learn probabilistic embeddings based on the variational information bottleneck (VIB) principle~\cite{vib1, vib2}.
    The soft contrastive loss is formulated as a probabilistic alternative to contrastive loss to handle the one-to-many mapping.
    With the HIB objectives, probabilistic cross-modal embeddings~\cite{pcme} have been studied to learn joint embeddings between images and captions for one-to-many image-text retrieval.
    In contrast with \cite{hib, pcme} trained in a supervised manner using image-text/label pairs, we learn probabilistic representations by only self-supervision without any labels.
    To this end, we propose a novel stochastic contrastive loss that is suited for self-supervised learning to optimize the model according to data uncertainty.

\vspace{-10pt}
\paragraph{Uncertainty in Computer Vision}
    As a method for improving the interpretability and the robustness to input data of deep neural networks, uncertainty have been extensively studied for a long time~\cite{uncertainty1, uncertainty2, uncertainty3}.
    In general, uncertainty is categorized into two types by different sources: (1) epistemic uncertainty (\ie, model uncertainty) which indicates uncertainty in the model parameters and (2) aleatoric uncertainty (\ie, data uncertainty) that originated from the inherent noise of data.
    Epistemic uncertainty can be reduced by providing enough training data~\cite{uncertainty1, uncertainty4}, whereas aleatoric uncertainty cannot be eliminated with additional training data~\cite{uncertainty3}.
    In computer vision, uncertainty has been explored for various tasks such as semantic segmentation~\cite{uncertainty3, uncertainty4}, object detection~\cite{uncertainty-detection1, uncertainty-detection2}, person re-identification~\cite{uncertainty-reid}, and face recognition~\cite{uncertainty-face1, uncertainty-face2, uncertainty-face3}.
    Although some works~\cite{uncertainty3, uncertainty-face2, uncertainty-face3} have considered the aleatoric uncertainty, they cannot be directly applied for self-supervised learning due to the absence of label information.
    Specific to deep video understanding, uncertainty has been used for video instance segmentation~\cite{video-segmentation}, weakly-supervised temporal action localization~\cite{video-wtal}, and video future frame prediction~\cite{video-prediction}.
    They have mainly focused on the predictive uncertainty estimated from the output of the model.
    In this work, we explore the aleatoric uncertainty of videos for self-supervised video contrastive learning.
    
    \vspace{-5pt}
    
    \begin{figure}[t]
    \begin{center}
    %\fbox{\rule{0pt}{2in} \rule{0.9\linewidth}{0pt}}
       \includegraphics[width=0.93\linewidth]{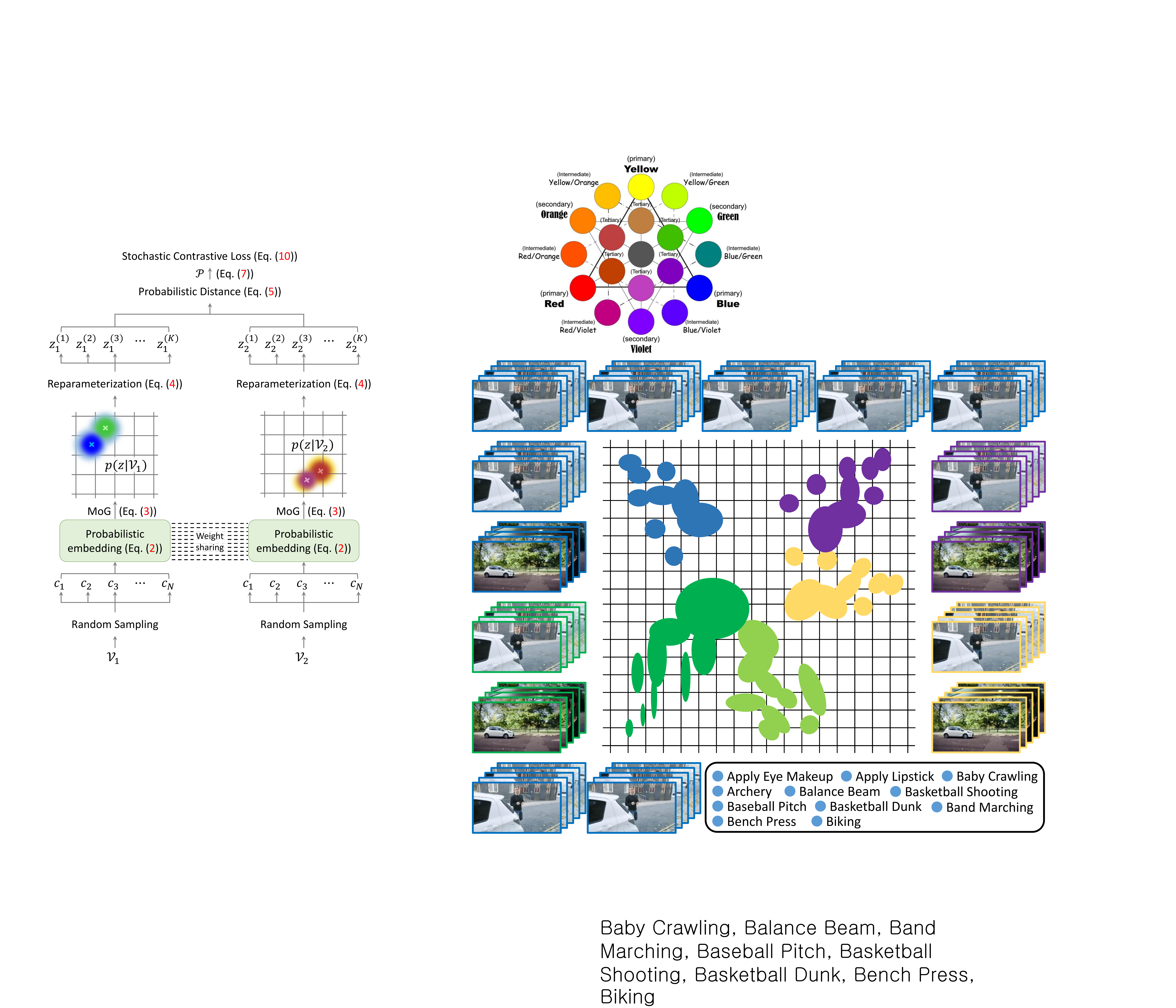}
    \end{center}
    \vspace{-15pt}
      \caption{\textbf{ProViCo} estimates the video distribution $p(z|\mathcal{V})$ as a Mixture of $N$ Gaussians with probabilistic clip embeddings. We construct the positive and negative pairs based on the probabilistic distance between two video distributions. The model learns probabilistic embedding network parameters and minimizes uncertainties of input videos through the stochastic contrastive loss.} 
    \label{fig:2}\vspace{-15pt}
    \end{figure}
    
\section{Method} \vspace{-5pt}
\subsection{Background and Motivation}\label{sec:motivation} \vspace{-5pt}
    Contrastive learning is a promising framework to learn video representations in a self-supervised manner.
    Given a fixed-length clip $q$ from a video $\mathcal{V}$, let $\lbrace{k_+}\rbrace$ be a set of positives sampled from the same video as $q$ (or augmented versions of $q$) and $\lbrace{k_-}\rbrace$ is a set of negatives sampled from other instances in a batch.
    The goal of contrastive learning is to maximize the similarities between $q$ and $\lbrace{k_+}\rbrace$ and minimize similarities between $q$ and $\lbrace{k_-}\rbrace$ through contrastive loss, such as the InfoNCE~\cite{infonce}:
    \begin{equation}\label{equ:infonce}
        \mathcal{L}_{\text{InfoNCE}} = -\log{\frac{\sum_{k\in{\lbrace{k_+}\rbrace}}{\exp({\text{sim}(q, k)/\tau})}}{\sum_{k\in{\lbrace{k_+, k_-}\rbrace}}{\exp({\text{sim}(q, k)/\tau})}}},
    \end{equation}
    where $\tau$ is a scaling temperature parameter and $\text{sim}(\cdot, \cdot)$ is a similarity function.
    The contrastive loss improves instance discrimination power by formulating relative distance between instances in the dynamic dictionaries~\cite{moco} instead of matching an input to a fixed target.
    However, as investigated in \cite{co-crl}, current approaches have focused on instance discrimination by treating each data sample as a ``class'', which makes the model neglect the semantic relations between different videos.
    We ascribe this to the risk of unstable matching arising from the lack of tools to measure the confidence of proximity between unlabeled instances during training.

    As a feasible solution, we propose the Probabilistic Representations for Video Contrastive Learning (ProViCo), in which videos are represented as probability distributions in a stochastic embedding space.
    In our framework, the uncertainty of videos is used as a key tool to measure the confidence of proximity between video distributions.
    Next we first describe the probabilistic embedding for video clips and extend it to the whole representation learning using a proposed stochastic contrastive loss.
    The overall procedure of ProViCo is illustrated in \figref{fig:2}.
 
\vspace{-5pt}   
\subsection{Probabilistic Video Embedding}\vspace{-5pt}
    Given a video $\mathcal{V}$, let $\lbrace{c_1,...,c_N}\rbrace$ be a set of clips sampled from $\mathcal{V}$, and $v_{c_n} = f_{\theta}(c_n)$ represents the output of the backbone network (\ie, encoder) parameterized by $\theta$.
    We formulate a probability distribution $p(z|c_n)$ for a clip $c_n$ as a normal distribution with a mean vector and a diagonal covariance matrix in a stochastic embedding space $\mathbb{R}^{D}$:
    \begin{equation}
        p(z|c_n) \sim \mathcal{N}(g_{\mu}(v_{c_n}), \text{diag}(g_{\sigma}(v_{c_n}))),
    \end{equation}
    where $g_{\mu}$ is a fully-connected (FC) layer followed by LayerNorm~\cite{layernorm} and $\ell_2$ normalization, and $g_{\sigma}$ is a separate FC layer without any normalization, following \cite{pcme}.
    With $N$ clip distributions, we represent the whole video distribution $p(z|\mathcal{V})$ as a Mixture of Gaussians~\cite{hib} such that
    \begin{equation}
        p(z|\mathcal{V}) \sim \sum_{n=1}^{N} \mathcal{N}(z; g_{\mu}(v_{c_n}), \text{diag}(g_{\sigma}(v_{c_n}))).
    \end{equation}
    From $p(z|\mathcal{V})$, we sample $K$ embeddings $\lbrace z^{(1)},...,z^{(K)}\rbrace \overset{\text{iid}}{\sim} p(z|\mathcal{V})$, that represent ``self-augmented" versions of the video representation.
    Specifically, we use reparameterization trick~\cite{reparameterization} for stable training, such that 
    \begin{equation}\label{equ:sampling}
        z^{(k)} = \sigma(\mathcal{V})\cdot \epsilon^{(k)} + \mu(\mathcal{V}),
    \end{equation}
    where $\mu(\mathcal{V})$, $\sigma(\mathcal{V})$ are the mean and the standard deviation of $p(z|\mathcal{V})$, and $\lbrace \epsilon^{(1)},...,\epsilon^{(K)} \rbrace$ are sampled iid from the $D$-dimensional unit Gaussian distribution.
  
 \vspace{-5pt}  
\subsection{Mining Positive and Negative Pairs}\label{sec:positive} \vspace{-5pt}
    Contrary to the deterministic representation methods~\cite{feichtenhofer21, video-moco}, we construct the positive and negative pairs based on the probabilistic distance that contains the uncertainty of embedded distributions.
    Specifically, given a embedding pair $z_i^{(k)} \sim p(z|\mathcal{V}_i)$ and $z_j^{(k')} \sim p(z|\mathcal{V}_j)$ sampled from $i$-th and $j$-th video distributions in a batch, we define the distance between two embeddings as Bhattacharyya distance~\cite{distance}:
    \begin{equation}\label{equ:distance}
        \begin{aligned}
            \text{dist}(z_i^{(k)}, z_j^{(k')}) = & \frac{1}{4} (\log{(\frac{1}{4}(\frac{\sigma_i^{2}}{\sigma_j^{2}} + \frac{\sigma_j^{2}}{\sigma_i^{2}} + 2))} \\
                        &+ \lambda\cdot\frac{(z_i^{(k)} \unaryminus z_j^{(k')})^\top (z_i^{(k)} \unaryminus z_j^{(k')})}{\sigma_i^{2} + \sigma_j^{2}}), % {\sigma_{i}^{2}\sigma_{j}^{2}},
        \end{aligned}
    \end{equation}
    where $\sigma_{i}^{2}$, $\sigma_{j}^{2}$ are variances for the $i$-th and $j$-th video distributions that represents the uncertainty of each video, and $\lambda$ is a scaling factor according to the dimension of the embedding space.
    The distance between two video distributions can be factorized via Monte-Carlo estimation:
    \begin{equation}
        \text{dist}(\mathcal{V}_i, \mathcal{V}_j) \approx \frac{1}{K^2} \sum_{k}^{K} \sum_{k'}^{K} \text{dist}(z_i^{(k)}, z_j^{(k')}).
    \end{equation}
    
    The positive pairs $\mathcal{P}$ are defined as all video pairs closer than threshold distance $\tau$, such that
    \begin{equation}\label{equ:positives}
        \mathcal{P} = \lbrace (\mathcal{V}_i , \mathcal{V}_j) \: | \: \text{dist}(\mathcal{V}_i , \mathcal{V}_j) < \tau \;\text{or}\; i=j\rbrace.
    \end{equation}
    Since we regard each embedding $z_i^{(k)}$ as a self-augmented version of $\mathcal{V}_{i}$, we also set a pair $(\mathcal{V}_i, \mathcal{V}_i)$ as a positive.
    The negative pairs are then the complement of the positive pairs (\ie, $\bar{\mathcal{P}}$).
    By defining positive and negative pairs based on the probabilistic distance, we can construct the confident sample pairs considering the uncertainty of videos.
    
\vspace{-5pt}
\subsection{Stochastic Contrastive Loss} \vspace{-5pt}
    As an alternative to conventional contrastive objectives such as \eqref{equ:infonce}, we introduce a stochastic contrastive loss to discriminate positive and negative pairs, and minimize the uncertainty of videos, simultaneously.
    The stochastic contrastive loss incorporates the uncertainty of each video into the soft contrastive loss~\cite{hib} that transformed contrastive loss for probabilistic embeddings.
    For a pair of videos $(\mathcal{V}_i, \mathcal{V}_j)$, the soft contrastive loss is formulated by
    \begin{equation}\label{equ:soft-contrastive}
        \mathcal{L}_{\text{soft}}(\mathcal{V}_i, \mathcal{V}_j) = \begin{cases}
                            \unaryminus \log{p(m|\mathcal{V}_i, \mathcal{V}_j}) 
                            & \text{if}\: (\mathcal{V}_i, \mathcal{V}_j) \in \mathcal{P} \\
                            \unaryminus \log{(1 \unaryminus p(m|\mathcal{V}_i, \mathcal{V}_j}))  
                            &  \text{otherwise}
                            \end{cases},
    \end{equation}
    where $p(m|\mathcal{V}_i, \mathcal{V}_j)$ is the match probability~\cite{hib, pcme} with the sigmoid function $s(\cdot)$ and learnable scalars $(a, b)$:
    \begin{equation}\label{equ:matchprob}
        p(m|\mathcal{V}_i, \mathcal{V}_j) = \frac{1}{K^2}\sum_k^K \sum_{k'}^K s(-a||z_i^{(k)} - z_j^{(k')}||_2 + b).
    \end{equation}
    
    Finally, the stochastic contrastive loss between $\mathcal{V}_i$ and $\mathcal{V}_j$ is defined as:
    \begin{equation}\label{equ:stochastic-contrastive-loss}
        \mathcal{L}_{\text{stoc}}(\mathcal{V}_i, \mathcal{V}_j) = \frac{1}{4\sigma_{i}^{2}\sigma_{j}^{2}}{\mathcal{L}_{\text{soft}}(\mathcal{V}_i, \mathcal{V}_j)}
                     + \frac{1}{2}(\log{\sigma_{i}^{2}} + \log{\sigma_{j}^{2}}),
    \end{equation}
    where the first term is for the instance discrimination between the probabilistic embeddings obtained with the model and the seconde term is a regularization term to prevent the model from predicting infinite uncertainty for all videos.
    Two terms complement each other to control the contribution of unreliable pairs and uncertainties.
    More concretely, the probabilistic distance in \eqref{equ:distance} is decreased for the video pair with substantial uncertainties, such that improper positive pairs can be constructed.
    However, high uncertainty (\ie, large $\sigma_i^2\sigma_j^2$) attenuates the contribution of the first term in the stochastic contrastive loss, penalizing unreliable pairs, and exaggerates the second term that reduces uncertainties.
    For pairs with low uncertainty, the model will focus on instance discrimination, as the first term have the larger contribution.
    These properties of the stochastic contrastive loss make the model robust to noisy videos.

\vspace{-5pt}   
\subsection{Total Objectives} \vspace{-5pt}
    We employ the additional KL regularization term between the video distribution and the unit Gaussian prior $\mathcal{N}(0, I)$ to prevent the predicted variance from collapse to zero, following \cite{pcme}:
    \begin{equation}
        \begin{aligned}
            \mathcal{L}_{\text{KL}}(\mathcal{V}_i, \mathcal{V}_j) = \text{KL}(p(z_i| & \mathcal{V}_i)|| \mathcal{N}(0, I)) \\ 
            &  + \text{KL}(p(z_j|\mathcal{V}_j)||\mathcal{N}(0, I)),
        \end{aligned}
    \end{equation}
    Therefore, the overall objective for ProViCo is a weighted sum of all loss functions defined as:
    \begin{equation}\label{equ:final-loss}
        \mathcal{L}_{\text{ProViCo}} = \mathcal{L}_{\text{stoc}} + \beta \cdot \mathcal{L}_{\text{KL}},
    \end{equation}
    where $\mathcal{L}_{\text{stoc}}$ and $\mathcal{L}_{\text{KL}}$ are the sum of each term for all pairs in a batch, and $\beta$ controls the trade-off between two terms.

    \begin{figure}[t]
    \begin{center}
       \includegraphics[width=0.95\linewidth]{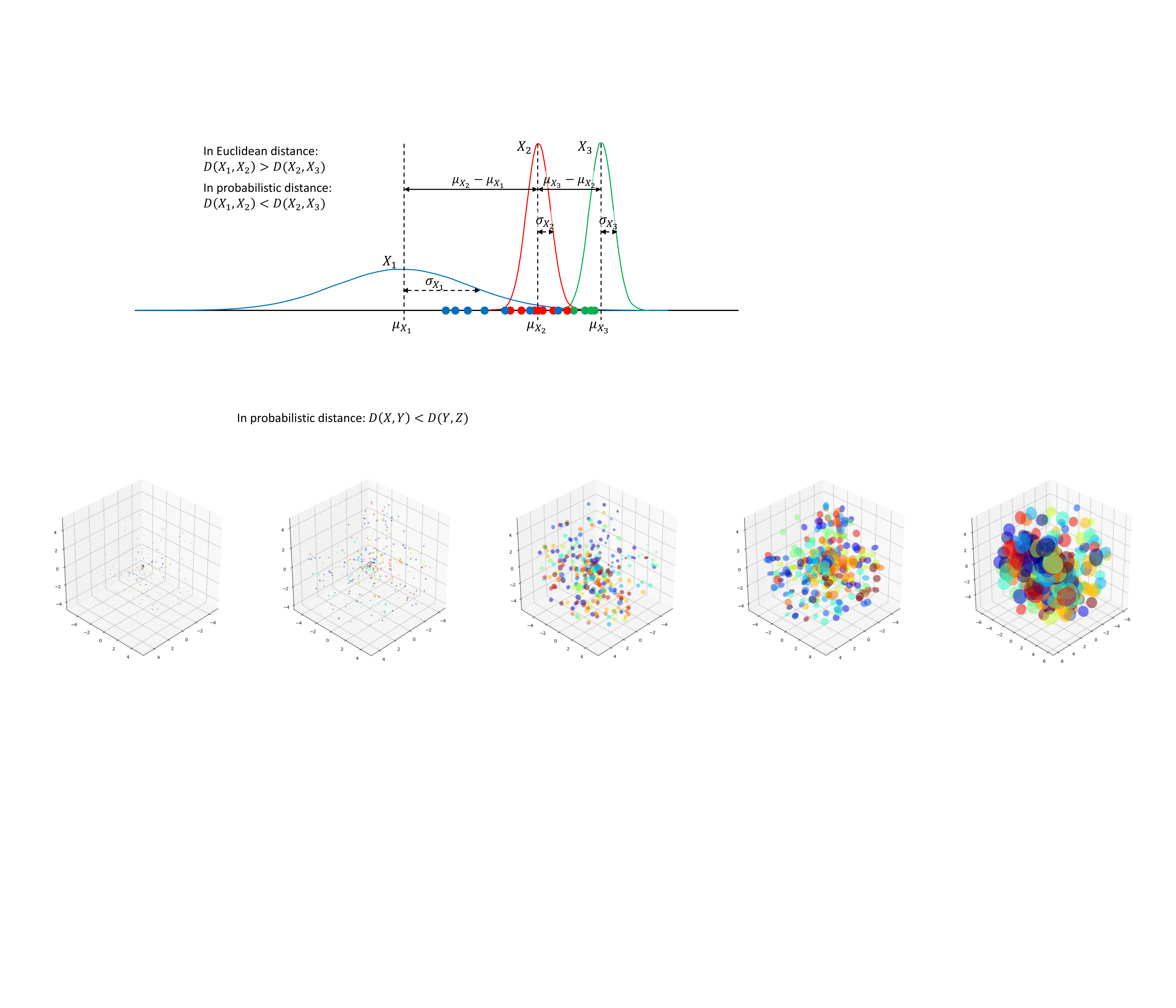}
    \end{center}
    \vspace{-15pt}
      \caption{\textbf{1-dimensional toy example for the distance between probability distributions.} The Euclidean distance between the mean values without considering variances cannot represent the probabilistic similarity between probability distributions.} 
      \vspace{-7pt}
    \label{fig:3}
    \end{figure}

\vspace{-5pt}  
\subsection{Rethinking Objectives with Distance} \vspace{-5pt}
    In our framework, the model learns to achieve two main objectives: instance discrimination in a stochastic embedding space and uncertainty minimization for input videos.
    For robust instance discrimination learning, we argue that the semantic relations beyond the instances are to be examined for mining positive and negative pairs.
    Although the conventional distance metric (\eg Euclidean distance or cosine similarity) can measure the similarity between instances using the mean of distributions, it is inadequate to represent the probabilistic similarity due to the variance of each distribution.
    For example, as illustrated in \figref{fig:3}, the higher similarity between probability distributions is not always guaranteed by the smaller Euclidean distance.
    From this observation, we determine the positive and negative pairs based on the probability distance between video distributions, as described in \secref{sec:positive}.
    On the other side, directly optimizing the probability distance to discriminate semantic instances may lead to unexpected results.
    Namely, minimizing the probability distance between positive pairs without any constraints can lead to an increase in the uncertainty of videos, decreasing the discrimination power of learned representations.
    To address this issue, our stochastic contrastive loss is designed to optimize the indirect probabilistic distance incorporating uncertainty and soft contrastive loss, which utilizes Euclidean distance.
    With such carefully designed training objectives, the model accomplish both discrimination of semantic instances and uncertainty minimization. 
    
    \section{Experiments}\vspace{-5pt}
\subsection{Datasets}
\vspace{-5pt}
\noindent\textbf{Kinetics-400~\cite{kinetics} (K400)}
    dataset contains $\sim$240k training videos of 400 human action categories.
    The test set consists of $\sim$38k videos, with about 100 videos for each class.
    We use the whole training videos to obtain initial parameters by pretraining the network using the proposed method.
    We measure action recognition performance on the test set using linear evaluation.

\noindent\textbf{UCF101~\cite{ucf101}}
    dataset consists of $\sim$13k videos of 101 action categories.
    To compare previous works~\cite{video-moco, mfo}, we perform pretraining on train split 1 and evaluate action recognition on test split 1 using two evaluation protocols (described in \secref{sec:setup}).
    The video retrieval performance is evaluated on test split 1 using nearest-neighbors.

\noindent\textbf{HMDB51~\cite{hmdb51}} 
    is a relatively small dataset, containing $\sim$7k videos of 51 action categories.
    Among the three splits, we use train split 1 for finetuning and measure action recognition and video retrieval performance on test split 1.
    
 \begin{table}
    \begin{center}
    \small
    \begin{tabular}{
    >{\raggedright}m{0.26\linewidth}>{\centering}m{0.35\linewidth}>{\centering}m{0.16\linewidth}}
        \hlinewd{0.8pt}
        Model                           & Backbone (\# params)  & Acc. (\%)  \tabularnewline
        \hline \hline
        \rowcolor{light_gray}
        Supervised                      &   R3D-50 (31.8M)  &   74.7    \tabularnewline
        Supervised                      &   R3D-18 (20.2M)  &   68.9    \tabularnewline
        Supervised                      &   R(2+1)D (15.4M)  &   71.7    \tabularnewline    \hline
        SeCo~\cite{seco}                &  ResNet-50 (23.0M)    &   61.9   \tabularnewline
        \rowcolor{light_gray}
        CVRL~\cite{cvrl}                &  R3D-50 (31.8M)    &   66.1   \tabularnewline
        \rowcolor{light_gray}
        CVRL~\cite{cvrl}                &   R3D-101 (51.4M)  &   67.6   \tabularnewline
        \rowcolor{light_gray}
        $\rho$BYOL~\cite{feichtenhofer21}     &  R3D-50 (31.8M)    &   68.3    \tabularnewline
        $\rho$MoCo~\cite{feichtenhofer21}     &  R(2+1)D (15.4M)    &   57.2    \tabularnewline
        % VideoMoCo~\cite{video-moco}     &  R3D-18 (20.2M)    &   74.1    \tabularnewline
        % VideoMoCo~\cite{video-moco}     &  R(2+1)D (15.4M)   &   78.7      \tabularnewline
        \rowcolor{light_gray}
        CORP~\cite{corp}                &  R3D-50 (31.8M)    &   66.3   \tabularnewline
        \rowcolor{light-green}
        \textbf{ProViCo (Ours)}                   &  R3D-18 (20.2M)    &   62.8       \tabularnewline
        \rowcolor{light-green}
        \textbf{ProViCo (Ours)}                   &   R(2+1)D (15.4M)  &    65.7   \tabularnewline
        \hlinewd{0.8pt}
        \end{tabular}\vspace{-17pt}
    \end{center}
    \caption{Linear evaluation for action recognition on the Kinetics-400~\cite{kinetics} dataset corresponding to the backbone networks. The models are pretrained on Kinetics-400. We shaded the considerably different experimental settings in terms of the backbone network, frame resolution, and batch size.
    }\label{tab:kinetics-linear}\vspace{-10pt}
    \end{table}

\vspace{-5pt}
\subsection{Experimental Setup}\label{sec:setup}
\vspace{-13pt}
\paragraph{Pretraining}
    For pretraining, we randomly sample two 16-frame clip with the temporal stride of 1 from each video.
    All frames in each clip are fixed to a size of $112 \times 112$ by random cropping.
    The backbone networks are trained for $200$ epochs with a mini-batch size of $96$.
    By using a half-period cosine learning rate scheduler~\cite{cosine-scheduler}, we warm-up the learning rate in the first 20 epochs from an initial learning rate of $10^{-4}$ with Adam optimizer~\cite{adam}.
    We set the scaling factor of the probabilistic distance $\lambda$ in \eqref{equ:distance} to $1/4D$ according to the embedding space size $D$ and threshold distance $\tau$ in \eqref{equ:positives} to $0.15$.
    The KL-divergence hyperparameter $\beta$ is set to $10^{-4}$ and the number of embeddings $K$ is set to $10$ throughout the experiments.

    \begin{table*}
    \begin{center}
    \small
    \begin{tabular}{
    >{\centering}m{0.04\linewidth}>{\raggedright}m{0.13\linewidth}>{\centering}m{0.16\linewidth}>{\centering}m{0.10\linewidth}>{\centering}m{0.08\linewidth}>{\centering}m{0.10\linewidth}>{\centering}m{0.055\linewidth}
    >{\centering}m{0.05\linewidth}>{\centering}m{0.05\linewidth}}
        \hlinewd{0.8pt}
        Type & Model                           & Backbone (\# params)& Input size  &   Batch size  & Pretrain data &     Finetune    &    UCF  &   HMDB  \tabularnewline
        \hline \hline
        % MemDPC~\cite{memdpc}                &  R2D3D-34 (?M)    &  $224 \times 224$  &   K400    &   \cmark  &   86.1    & 54.5  \tabularnewline
        % PRP~\cite{prp}             &  R(2+1)D (15.4M)    &  $224 \times 224$  & - &   UCF101    &   \xmark  &   32.1    &   -   \tabularnewline    % check input size
        (\romannumeral 2) & MFO~\cite{mfo}             &  R3D-18 (20.2M)    &  $112 \times 112$  & 256 &   K400    &   \xmark  &   63.2    &   33.4   \tabularnewline
        \rowcolor{light_gray}
        (\romannumeral 2) & CATE~\cite{cate}             &  R3D-50 (31.7M)    &  $224 \times 224$  & 1024 &   K400    &   \xmark  &   84.3    &   53.6   \tabularnewline
        \rowcolor{light_gray}
        (\romannumeral 3) & CORP~\cite{corp}             &  R3D-50 (31.7M)    &  $224 \times 224$  & 512 &   K400    &   \xmark  &   90.2    &   58.7   \tabularnewline
        \rowcolor{light-green}
        (\romannumeral 2) & \textbf{ProViCo (Ours)}                   &  R3D-18 (20.2M)    &  $112 \times 112$  & 96 &   K400    &   \xmark  &   82.9  &   52.2    \tabularnewline
        \rowcolor{light-green}
        (\romannumeral 2) & \textbf{ProViCo (Ours)}                   &   R(2+1)D (15.4M)  &  $  112 \times 112$  & 96 &   K400    &   \xmark  &   84.1      &   53.5    \tabularnewline
        \hlinewd{0.8pt}
        (\romannumeral 1) & DSM~\cite{dsm}                &  I3D (25.0M)    &  $224 \times 224$  & 128 &   K400    &   \cmark  &   74.8    & 52.5  \tabularnewline
        \rowcolor{light_gray}
        (\romannumeral 2) & CVRL~\cite{cvrl}                &  R3D-50 (31.7M)    &  $224 \times 224$  & 1024 &   K400    &   \cmark  &   92.2    & 66.7  \tabularnewline
        (\romannumeral 2) & VideoMoCo~\cite{video-moco}     &  R(2+1)D (15.4M)   &  $112 \times 112$  & 128 &   K400    &   \cmark  &   78.7    &   49.2   \tabularnewline
        (\romannumeral 2) & MFO~\cite{mfo}                  &  R3D-18 (20.2M)    &  $112 \times 112$  & 256 &   K400    &   \cmark  &   79.1    &   47.6   \tabularnewline
        \rowcolor{light_gray}
        (\romannumeral 2) & CATE~\cite{cate}                  &  R3D-50 (31.7M)    &  $  128 \times 128$  & 1024 &   K400    &   \cmark  &   88.4    &   61.9   \tabularnewline
        \rowcolor{light_gray}
        (\romannumeral 2) & MCN$^\dagger$~\cite{mcn}                  &  R3D-18 (20.2M)    &  $  128 \times 128$  & 80 &   K400    &   \cmark  &   89.7    &   59.3   \tabularnewline
        (\romannumeral 1) & TCLR~\cite{tclr}             &  R(2+1)D (15.4M)    &  $112 \times 112$  & 40 &   K400    &   \cmark  &   84.3    &   54.2   \tabularnewline
        (\romannumeral 3) & TEC~\cite{tec}             &  R(2+1)D (15.4M)    &  $112 \times 112$  & 192 &   K400    &   \cmark  &   87.1    &   59.8   \tabularnewline
        \rowcolor{light-green}
        (\romannumeral 2) & \textbf{ProViCo (Ours)}                   &  R3D-18 (20.2M)    &  $112 \times 112$  & 96 &   K400    &   \cmark  &   85.6    &    58.4   \tabularnewline
        \rowcolor{light-green}
        (\romannumeral 2) & \textbf{ProViCo (Ours)}                   &   R(2+1)D (15.4M)  &  $112 \times 112$  & 96 &   K400    &   \cmark  &   87.2   &  59.4     \tabularnewline
        \hlinewd{0.8pt}
        \rowcolor{light_gray}
        (\romannumeral 1) & VCP~\cite{vcp}                  &  C3D (34.8M)       &  $  112 \times 112$  & - &   UCF101    &   \cmark  &   68.5    &  32.5    \tabularnewline
        (\romannumeral 1) & PRP~\cite{prp}                  &  R(2+1)D (15.4M)       &  $  112 \times 112$  & - &   UCF101    &   \cmark  &   72.1    &  35.0    \tabularnewline
        (\romannumeral 1) & RTT~\cite{rtt}                  &  R(2+1)D (15.4M)       &  $  112 \times 112$  & 256 &   UCF101    &   \cmark  &   81.6    &  46.4    \tabularnewline
        \rowcolor{light_gray}
        (\romannumeral 1) & DSM~\cite{dsm}                &  C3D (34.8M)    &  $  112 \times 112$  & 128 &   UCF101    &   \cmark  &   70.3    & 40.5  \tabularnewline
        (\romannumeral 2) & MFO~\cite{mfo}                  &  R3D-18 (20.2M)    &  $  112 \times 112$  & 256 &   UCF101    &   \cmark  &   76.2    &   41.1   \tabularnewline
        \rowcolor{light_gray}
        (\romannumeral 2) & MCN$^\dagger$~\cite{mcn}                  &  R3D-18 (20.2M)    &  $  128 \times 128$  & 80 &   UCF/HMDB    &   \cmark  &   85.4    &   54.8   \tabularnewline
        \rowcolor{light_gray}
        (\romannumeral 3) & CORP~\cite{corp}             &  R3D-50 (31.7M)    &  $224 \times 224$  & 512 &   UCF/HMDB    &   \cmark  &   93.5    &   68.0   \tabularnewline
        (\romannumeral 2) & TCLR~\cite{tclr}             &  R(2+1)D (15.4M)    &  $112 \times 112$  & 40 &   UCF101    &   \cmark  &   82.8    &   53.6   \tabularnewline
        (\romannumeral 3) & TEC~\cite{tec}             &  R(2+1)D (15.4M)    &  $112 \times 112$  & 192 &   UCF101    &   \cmark  &   85.2    &   56.9   \tabularnewline
        \rowcolor{light-green}
        (\romannumeral 2) & \textbf{ProViCo (Ours)}                   &  R3D-18 (20.2M)    &  $  112 \times 112$  & 96 &   UCF101    &   \cmark  &   83.7  &   57.1    \tabularnewline
        \rowcolor{light-green}
        (\romannumeral 2) & \textbf{ProViCo (Ours)}                   &   R(2+1)D (15.4M)  &  $  112 \times 112$  & 96 &   UCF101    &   \cmark  &     86.1    &    58.0   \tabularnewline
        \rowcolor{light-green}
        (\romannumeral 2) & \textbf{ProViCo (Ours)}                   &   R3D-50 (31.7M)  &  $  224 \times 224$  & 512 &   UCF101    &   \cmark  &     94.6    &    68.2   \tabularnewline
        \hlinewd{0.8pt}
        \end{tabular}
    \end{center}
    \vspace{-17pt}
    \caption{Action recognition performance on UCF101~\cite{ucf101} and HMDB51~\cite{hmdb51} dataset corresponding to the pretrained dataset and backbone networks. \textbf{Finetune} \cmark\;means the whole networks are finetuned end-to-end, while \xmark\;means the backbone network is fixed and the linear classifier is updated only. We shaded the considerably different experimental settings in terms of the backbone network, frame resolution, and batch size. $\dagger$ denotes that additional residual views are used~\cite{mcn}.
    }\label{tab:ucf-action}\vspace{-10pt}
    \end{table*}
    
\vspace{-10pt}
\paragraph{Backbone networks}
    To provide a fair comparison, we employ two popular 3D networks as backbone networks: R3D-18~\cite{r3d, r3d2} and R(2+1)D-18~\cite{r(2+1)d}.
    These architectures have model parameters of 20.2M and 15.4M, respectively, which are much lighter compared to models such as R3D-50 or R3D-101.
    For action recognition, we evaluate the performance using both backbone networks.
    For video retrieval and ablation studies, we report only the results of R3D-18.

\vspace{-10pt}
\paragraph{Evaluation protocols}
    We evaluate the proposed method for action recognition and video retrieval tasks.
    Following previous works~\cite{feichtenhofer21, cvrl}, we adopt two evaluation protocols to verify the learned video representations:
    (\romannumeral 1) \textit{Linear evaluation} provides a straightforward evaluation for learned representations by fixing all the parameters in the backbone network and finetuning only the fully-connected (FC) layers.
    (\romannumeral 2) \textit{Finetuning} updates parameters in both the pre-trained backbone and the additional FC layers.
    For action recognition, we pretrain the backbone network on the Kinetics-400~\cite{kinetics} and UCF101~\cite{ucf101} datasets, respectively.
    We report the top-1 accuracy evaluated on the Kinetics-400 and UCF101 datasets using two evaluation protocols.
    In addition, we evaluate the performance on the HMDB51~\cite{hmdb51} dataset using finetuning.
    For video retrieval, we measure top-1, 5, 10, and 20 accuracies using nearest-neighbors without additional training and compare with state-of-the-art methods on UCF101 and HMDB51 datasets.

\vspace{-6pt}
\subsection{Action Recognition}
\vspace{-5pt}
    We first compare the action recognition performance of ProViCo with state-of-the art methods.
    In our framework, the action of the video is predicted by averaging the output probabilities of the classifier for all embeddings sampled in \eqref{equ:sampling}.
    We observe that the performance is significantly affected by the architecture of backbone networks, the video frame resolution, and the batch size used during training.
    Since we set the minimum level of these components, we mainly compare the results with similar conditions.
  
    \begin{table*}[t]
    \centering
    \small
    \resizebox{0.95\linewidth}{!}{
        \begin{tabular}{lcc|cccc|cccc}
        \hlinewd{0.8pt}
        \multirow{2}{*}{Method}    & \multirow{2}{*}{Backbone (\# params)}   & \multirow{2}{*}{Pretrain}   & \multicolumn{4}{c|}{UCF101} &  \multicolumn{4}{c}{HMDB51} \tabularnewline
        \cline{4-7} \cline{8-11}
             &   &  & R@1 & R@5 & R@10 & R@20 & R@1 & R@5 & R@10 & R@20\tabularnewline
        \hline \hline
        SpeedNet~\cite{speed-net}           &    S3D-G (9.1M)   &   K400     &  13.0  &     28.1     &  37.5  & 49.5    &   -   &   -   &   -   &   -   \tabularnewline
        MFO~\cite{mfo}          &    R3D-18 (20.2M)  &   K400     &  41.5  &    60.6     &  71.2  & 80.1     &   20.7   &   40.8   &   55.2   &  68.3    \tabularnewline
        CATE~\cite{cate}          &    R3D-50 (31.7M)  &   K400     &  54.9  &    68.3     &  75.1  & 82.3     &   33.0   &   56.8   &   69.4   &  82.1    \tabularnewline
        TEC~\cite{tec}          &    R3D-18 (20.2M)  &   K400     &  66.9  &    \textbf{83.1}     &  88.8  & 93.3     &   36.4   &   \textbf{64.1}   &   74.1   &  83.8    \tabularnewline
        \rowcolor{light-green}
        \textbf{ProViCo (Ours)}        & R3D-18 (20.2M)    & K400    &\textbf{67.6}   &81.4 &\textbf{90.1}  &\textbf{94.7}   &\textbf{40.1}   & 60.6 &  \textbf{75.2}  &\textbf{85.2}  \tabularnewline \hlinewd{0.8pt}
        VCP~\cite{vcp}             &    R3D-18 (20.2M)   &   UCF101     &  18.6  &    33.6     &  42.5  & 53.5    &   7.6   &   24.4   &   36.3   &   53.6    \tabularnewline
        Pace~\cite{pp}             &    R3D-18 (20.2M)  &   UCF101     &  23.8  &    38.1     &  46.4  & 56.6  &   9.6   &   26.9   &   41.1   &   56.1    \tabularnewline
        PRP~\cite{prp}             &    R3D-18 (20.2M)   &   UCF101     &  22.8  &    38.5     &  46.7  & 55.2     &   8.2   &   25.8   &   38.5   &   53.3    \tabularnewline
        DSM~\cite{dsm}             &    I3D (25.0M)  &   UCF101     &  17.4  &    35.2     &  45.3  & 57.8        &   7.6   &   23.3   &   36.5   &   52.5    \tabularnewline
        STS~\cite{sts}             &    R3D-18 (20.2M)  &   UCF101     &  38.3  &    59.9     &  68.9  & 77.2     &   18.0   &   37.2   &   50.7   &  64.8    \tabularnewline
        MFO~\cite{mfo}          &    R3D-18 (20.2M)  &   UCF101     &  39.6  &    57.6     &  69.2  & 78.0     &   18.8   &   39.2   &   51.0   &  63.7    \tabularnewline
        MCN~\cite{mcn}$^\dagger$          &    R3D-18 (20.2M)  &   UCF101     &  53.8  &    70.2     &  78.3  & 83.4     &   24.1   &   46.8   &   59.7   &  74.2    \tabularnewline
        TCLR~\cite{tclr}          &    R3D-18 (20.2M)  &   UCF101     &  56.2  &    72.2     &  79.0  & 85.3     &   22.8   &   45.4   &   57.8   &  73.1    \tabularnewline
        TEC~\cite{tec}          &    R3D-18 (20.2M)  &   UCF101     &  62.5  &    \textbf{78.4}     &  84.1  & 88.8     &   32.0   &   \textbf{60.8}   &   72.2   &  81.7    \tabularnewline
        \rowcolor{light-green}
        \textbf{ProViCo (Ours)}        & R3D-18 (20.2M)   & UCF101  &\textbf{63.8}   &  75.1 &\textbf{84.8}  &\textbf{89.2}   &\textbf{35.9}   &    55.2 &\textbf{74.3}  &\textbf{81.8}  \tabularnewline
        \hlinewd{0.8pt}
        \end{tabular}
        }\vspace{-7pt}
    \caption{Performance comparisons for video retrieval task evaluated on UCF101~\cite{ucf101} and HMDB51~\cite{hmdb51} datasets. We report top-1, 5, 10, 20 accuracies. $\dagger$ denotes that additional residual views are used~\cite{mcn}.
    }\label{tab:other}\vspace{-10pt}
    \end{table*}
 
    \vspace{-10pt}
    \paragraph{Linear evaluation on K400}
    We report the linear evaluation results on Kinetics-400~\cite{kinetics} in \tabref{tab:kinetics-linear}.
    The first three rows represent the results of supervised learning for each backbone network, showing the significant performance gap according to the network.
    The comparison between our method and $\rho$MoCo~\cite{feichtenhofer21} shows that our probabilistic approach outperforms the deterministic approach by a large margin (8.5\% performance gain with the same backbone networks).
    Compared to methods (CVRL~\cite{cvrl}, $\rho$BYOL~\cite{feichtenhofer21} and CORP~\cite{corp}) using about twice as many network parameters as R(2+1)D, our method attains competitive performance, further reducing the gap between self-supervised and supervised learning.

    \vspace{-10pt}
    \paragraph{Linear evaluation on UCF101 and HMDB51}
    We provide the linear evaluation results on UCF101~\cite{ucf101} and HMDB51~\cite{hmdb51} datasets in the first block of \tabref{tab:ucf-action}.
    The models are pretrained on Kinetics-400 dataset and evaluated on each dataset using linear evaluation.
    The results show that our method outperforms MFO~\cite{mfo} with about $\times2.5$ smaller batch size, showing 19.7\% and 18.8\% performance improvements on UCF101 and HMDB51 datasets, respectively.
    In addition, our method provides comparable performance to CATE\cite{cate} and CORP~\cite{corp}, even though they used $\times2$ more parameters, $\times2$ larger frame resolution, and $\times5$ and $\times10$ larger batch size, respectively.
    
    \vspace{-10pt}
    \paragraph{Finetuning on UCF101 and HMDB51}
    In the second and third blocks of \tabref{tab:ucf-action}, we report the comparison results for finetuning evaluation protocol.
    While the methods in the second blocks are pretrained on the Kinetics-400, the methods in the third blocks are pretrained on the UCF101 which is more smaller dataset than Kinetics-400.
    We analyze the results by dividing previous approaches into three aspects:
    (\romannumeral 1) pretext tasks without contrastive learning~\cite{dsm, vcp, prp, rtt};
    (\romannumeral 2) contrastive learning~\cite{cvrl, video-moco, mfo, cate, mcn, tclr};
    (\romannumeral 3) contrastive learning with pretext tasks using additional classifier branches~\cite{corp, tec}.
    First of all, the results show that our method significantly outperforms methods of pretext tasks~\cite{dsm, vcp, prp, rtt} without contrastive learning on both pretraining datasets, regardless of the backbone network, the frame resolution, and the batch size.
    In comparison with contrastive learning approaches~\cite{cvrl, video-moco, mfo, cate, mcn, tclr}, our method achieves significant improvements, except for \cite{cvrl, cate, corp}, which use a much larger batch size and a considerably deeper architecture with higher computational requirements.
    While MCN~\cite{mcn} used additional residual views with RGB view, our method shows competitive performances and even outperforms when the network pretrained on UCF101.
    The results of TEC~\cite{tec} pretrained on Kinetics-400 show that the deterministic approach achieves a similar performance to our probabilistic approach by combining contrastive learning and pretext tasks that use additional parameters, as mentioned in \cite{moco}.
    However, our method demonstrates robustness to the small number of training videos in the results pretrained on UCF101, showing 0.9\% and 1.4\% performance degradation on each dataset, while 1.9\% and 2.9\% degradation in \cite{tec}.
    With the deeper architecture (\ie R3D-50) and a larger batch size, our method achieves state-of-the-art performance, outperforming CORP~\cite{corp} by 1.1\% and 0.2\% on UCF101 and HMDB51, respectively.

 \vspace{-5pt}
\subsection{Video Retrieval} \vspace{-5pt}
    \tabref{tab:other} presents the video retrieval performance on UCF101 and HMDB51 datasets according to the pretraining dataset.
    For the video retrieval, we compute nearest-neighbors using the match probability in~\eqref{equ:matchprob} between two videos with Monte-Carlo estimation, unlike prior works, which used cosine similarity.
    We also provide the results using cosine similarity in the supplementary material.
    Our model achieves significantly improved top-1 accuracy performance on overall experimental results regardless of the network architecture~\cite{cate} and input data~\cite{mcn}.
    The results also show the advantage of learning probabilistic video representations over utilizing contrastive learning to learn deterministic video representations, as in previous works~\cite{mfo, cate, mcn, tclr}.
    We ascribe these results to our probabilistic approach, which learns probabilistic distributions of data and utilizes hard positive pairs so that learned representations are more suitable for matching tasks.

    \begin{table}[]
    \centering
    \small
    \begin{tabular}{
    >{\raggedright}m{0.3\linewidth}
    >{\centering}m{0.1\linewidth}>{\centering}m{0.1\linewidth}
    >{\centering}m{0.1\linewidth}>{\centering}m{0.1\linewidth}
    }
    \hlinewd{0.8pt}
    \multicolumn{5}{l}{\textbf{KL-divergence hyperparameter} ($K = 10$, $N = 2$)}                                 \tabularnewline 
    % \multicolumn{5}{l}{\textbf{KL-divergence hyperparameter}}                       \\
    Parameter $\beta$              &         $10^{-5}$             &          $10^{-4}$            &         $10^{-3}$             &          $10^{-2}$            \tabularnewline
    \hline
    Acc. (\%)                  &            -          &           \textbf{83.7}           &          83.4            &          81.6            \tabularnewline \hlinewd{0.8pt}
    \multicolumn{5}{l}{\textbf{Number of sampled embeddings} ($\beta = 10^{-4}$, $N = 2$)}                              \tabularnewline  
    Parameter $K$                &          5            &         7             &          10            &         12             \tabularnewline
    \hline
    Acc. (\%)                     &          81.2            &          83.1            &           \textbf{83.7}           &          -            \tabularnewline \hlinewd{0.8pt}
    \end{tabular}\vspace{-7pt}
    \caption{Ablation studies for KL-divergence hyperparameter $\beta$ and the number of sampled embeddings $K$.}\label{tab:ablation}\vspace{-12pt}
    \end{table}

\begin{figure}[t]
        \centering
           {\includegraphics[width=0.99\linewidth]{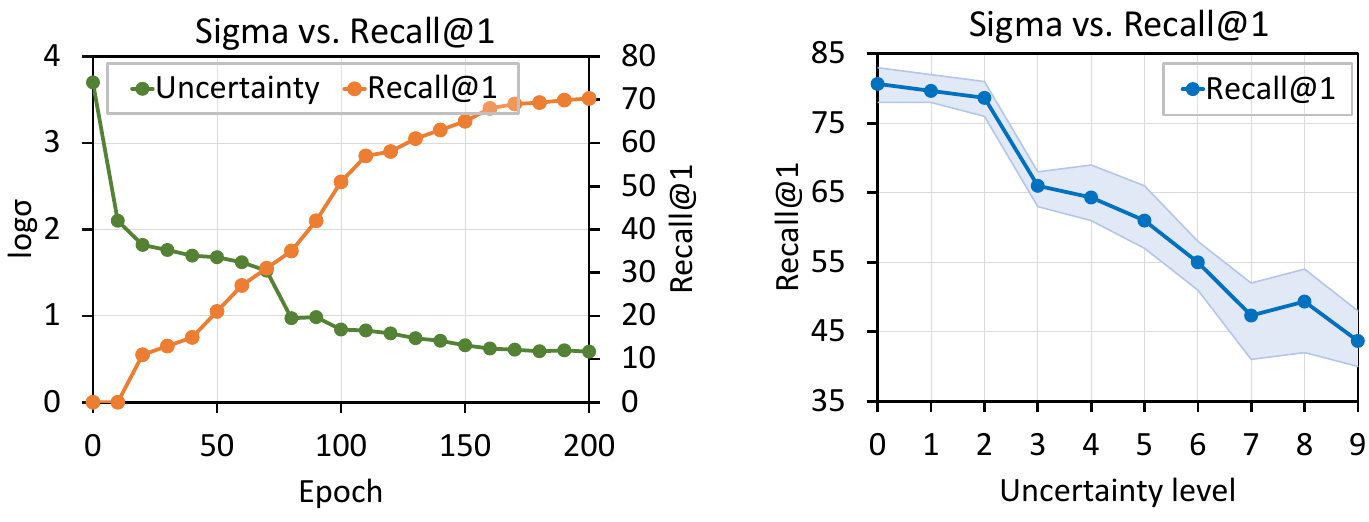}\label{fig:4}}\hfill     
        %   {\includegraphics[width=0.467\linewidth]{figure/fig4-b.pdf}\label{fig:4b}}\hfill
        \vspace{-10pt}
        \caption{(\textbf{Left}) \textbf{Uncertainty versus performance} during training. (\textbf{Right}) \textbf{Performance versus uncertainty level} for test set.
        }
    \label{fig:4}\vspace{-7pt}
    \end{figure}
    
\vspace{-5pt}
\subsection{Ablation Study and Analysis} \vspace{-5pt}
    To further validate and fully investigate the components of our method, we conduct ablation experiments for action recognition according to the value of the KL-divergence hyperparameter $\beta$ and the number of sampled embeddings $K$. %, and the number of clips $N$ used to estimate video distributions.
    In addition, we provide the uncertainty analysis to verify the impact of the uncertainty on representation learning.
    For all experiments in this section, we use R3D-18 as the backbone network and batch size is fixed to 96.
    Note that the backbone network is pretrained and evaluated on UCF101~\cite{ucf101} using finetuning for action recognition and nearest-neighbors for video retrieval.

\vspace{-10pt}
\paragraph{KL-divergence hyperparameter}
    We report the action recognition performance in accuracy according to the value of the hyperparameter $\beta$ in \eqref{equ:final-loss} to explore the effect of the KL-divergence regularization.
    As shown in the first block of \tabref{tab:ablation}, the maximum performance is obtained with $\beta=10^{-4}$.
    Formally, an increase in $\beta$ yields that the variance of the embedding is learned close to the unit variance, reducing the discriminability of distributions.
    On the contrary, when $\beta$ is too small ($10^{-5}$), the stochastic contrastive loss diverges as the variance approaches zero.

\vspace{-10pt}
\paragraph{Number of sampled embeddings}
    In the second block of \tabref{tab:ablation}, we report the performance according to the number of sampled embeddings $K$ in \eqref{equ:sampling}.
    At testing time, we use the same number of embeddings and average the output of the classifier to predict the class of the video.
    Since large numbers of embeddings reflect the entire distribution of the video via Monte-Carlo estimation, the performance increased as $K$ increases.
    However, a larger number of $K$ leads to more computational requirements.
    We choose $K = 10$ in consideration of computational costs.

\vspace{-10pt}
\paragraph{Uncertainty and performance}
    To analyze the correlation between the uncertainty and the discriminability of learned representations, we measure the inherent uncertainty of videos and report the video retrieval performance on UCF101 according to the training step.
    For every 10 epochs, we estimate the average uncertainty for all videos in the training set and compute the average of top-1 accuracy.
    As shown in the left side of \figref{fig:4}, the model learns to minimize uncertainty of videos, thereby improving the retrieval performance.
    Furthermore, we analyze the correlation between the performance and the uncertainty level by evaluating the retrieval performance on three test splits of UCF101.
    We divide the uncertainty measured for all videos into 10 uniform bins according to the uncertainty level and compute the average top-1 performance in each of the bins.
    As presented in the right side of \figref{fig:4}, results show the negative correlation between the uncertainty and the performance, indicating the performance drops as the uncertainty increases.
    The additional uncertainty analyses with visualization are presented in the supplementary material.
    
    \vspace{-7pt}
\section{Conclusion} \vspace{-7pt}
    We have introduced ProViCo that learns video representations in a self-supervised manner by estimating the distribution and the uncertainty of videos in a stochastic embedding space.
    The probabilistic framework provides a discriminative sample embedding without any spatiotemporal transformations, while not impairing the nature of the video (by artificial transformations).
    We constructed the positive and negative pairs based on the probabilistic distance to hold more semantically related candidates for robust contrastive learning without class annotations.
    The proposed stochastic contrastive loss enables not only the learning of video representations from reliable sample pairs by attenuating the impact of uncertain samples but also minimization of uncertainty from the inherent nature of the raw video.
    Extensive experiments showed that the probabilistic embedding can be a powerful alternative to the deterministic counterparts, achieving state-of-the-art performance.
    
    \vspace{-7pt}
    \section{Broader Impact} \vspace{-7pt}
    Self-supervised video representation learning is an appealing topic in computer vision with many downstream applications.
    A successful representation learning framework (such as that presented in this work) takes a significant step toward realizing these applications by alleviating the huge financial and environmental costs that would otherwise be necessary.
    To promote relative research, we discuss that there are possible directions for future work.
    ProViCo represents the probabilistic video distributions without a temporal encoding.
    Beyond the simple contrastive learning, pretext tasks to enforce the temporal encoding~\cite{corp, tec} may further improve the capacity of learned representations.
    In addition, while the improvement of ProViCo is consistently competitive on the Kinetics-400, UCF101, and HMDB51 benchmarks, all videos in these benchmark datasets are ``trimmed videos."
    It would be interesting to consider ``untrimmed videos" that randomly sampled clips can not be directly utilized for training due to severe background clutter, and their resultant high aleatoric uncertainty.
    We hope that our uncertainty-based approach will be a valuable foundation for video representation learning on large-scale untrimmed video datasets such as ActivityNet~\cite{activity-net}.

{\small
\bibliographystyle{ieee_fullname}
\bibliography{egbib}
}
    \clearpage
    \clearpage
    \newpage
    \section*{Appendix}
    \appendix
    
    \setcounter{figure}{0}
    \setcounter{table}{0}
    \setcounter{equation}{0}
    
        In this document, we include supplementary materials for ProVico.
    We first describe methodological details on ProViCo (\secref{sec:1}, \secref{sec:mining}) and provide implementation details (\secref{sec:3}) for downstream tasks.
    The additional experimental results (\secref{sec:4}), including ablation studies and qualitative results, are also presented to complement the main paper.
    Finally, observations and intuitive analyses to leverage the uncertainty for the future work are introduced in \secref{sec:5}.
    For all experiments in this document, we use R(2+1)D~\cite{r(2+1)d} as the backbone network. 
    
    \section{Uncertainty Computation on ProViCo}\label{sec:1}
    In ProViCo, the whole video distribution is esimated as a Mixture of Gaussians (MoG) with $N$ clip embeddings.
    Moreover, the variance predicted for each video represents the inherent uncertainty for the data.
    To compute the uncertainty of videos, we briefly provide the computations of the mean and variance of the whole video distribution from the clip distribution.
    Given a set of $N$ clips $\lbrace{c_n}\rbrace_{n=1}^{N}$ sampled from a video $\mathcal{V}$, let $f_{c_n}(z)$ be a probability density function (PDF) for the $n$-th clip with a mean vector $\mu_n \in \mathbb{R}^{D}$ and a diagonal components of covariance matrix $\sigma_n^2 \in \mathbb{R}^{D}$.
    The PDF of the MoG is represented by the averaged PDF of $N$ embeddings:
    \begin{equation}
        f_{\mathcal{V}}(z) = \frac{1}{N}\sum_{n}f_{c_n}(z).
    \end{equation}
    Then, the average vector $\mu_\mathcal{V}$ for the MoG is computed by the averaged mean vectors from clips:
    \begin{equation}
    \begin{aligned}
        \mu_{\mathcal{V}} &= \int z f_{\mathcal{V}}(z) dz   \\
                        & = \frac{1}{N} \sum_{n} \int z f_{c_n}(z) dz \\
                        & = \frac{1}{N} \sum_{n} \mu_{n}.
    \end{aligned}
    \end{equation}
    The variance $\sigma_{\mathcal{V}}^{2}$ for the MoG is derived as follows:
    \begin{equation}\label{equ:variance}
        \begin{aligned}
            \sigma_{\mathcal{V}}^2 &= \int z^2 f_{\mathcal{V}}(z) dz - \mu_\mathcal{V}^{2}\\
                                &= \frac{1}{N}\sum_n \int z^2 f_{c_n}(z) dz - \mu_\mathcal{V}^2\\
                                &= \frac{1}{N}\sum_n (\sigma_n^2 + \mu_n^2) - \mu_\mathcal{V}^2\\
                                &= \frac{1}{N}\sum_n (\sigma_n^2 + \mu_n^2) - (\frac{1}{N}\sum_n \mu_n)^2.
        \end{aligned}
    \end{equation}
    Finally, the geometric mean of the variance $\sigma_{\mathcal{V}}^2$ is used as the uncertainty of $\mathcal{V}$.
    
    \begin{figure}[t]
    \begin{center}
       \includegraphics[width=1\linewidth]{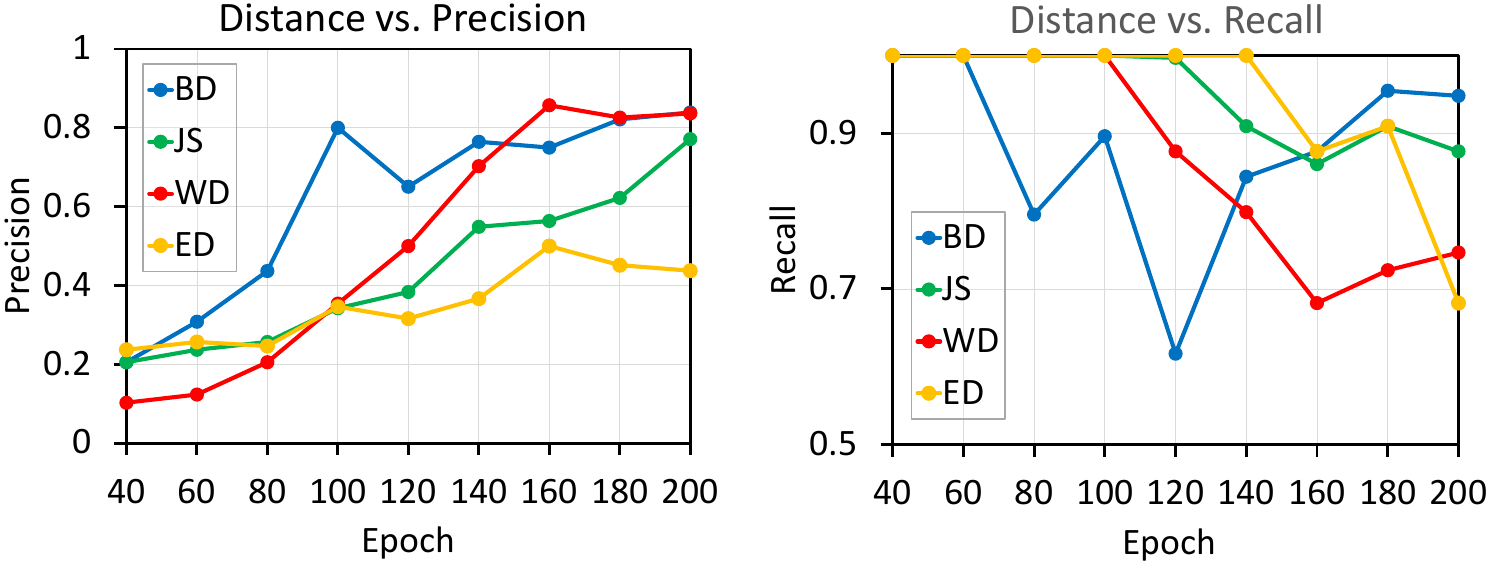}
    \end{center}
    \vspace{-15pt}
      \caption{\textbf{(Left) Distance versus precision} and \textbf{(Right) Distance versus recall} at every 20 epochs. Since we do not consider hard positives at the first 30 epochs, the precision and recall are computed from epoch 40. \textbf{BD}: Bhattacharyya distance, \textbf{JS}: Jensen-Shannon divergence, \textbf{WD}: Wasserstein distance, and \textbf{ED}: Elidean distance.}
    \label{fig:precision-recall}
    \end{figure}
    
    \section{Probabilistic Distance}\label{sec:mining}
    \subsection{Other Distance Metrics}
    In the main paper, Bhattacharyya distance is used to measure the probabilistic distance between two normal distributions.
    We extend the probabilistic distance to various formulations and analyze them in terms of positive and negative mining.
    To simplify each distance, we first define Euclidean distance\footnote{Strictly the square of Euclidean distance.} as $d(\cdot)$ between two sampled embeddings:
    \begin{equation}
        d(z_p^{(k)}, z_q^{(k')}) = (z_p^{(k)} \unaryminus z_q^{(k')})^{\top}(z_p^{(k)} \unaryminus z_q^{(k')}),
    \end{equation}
    where $z_{p}^{(k)}$ and $z_{q}^{(k')}$ are sampled from different distributions $p = \mathcal{N}(\mu_p, \sigma_p^2)$ and $q = \mathcal{N}(\mu_q, \sigma_q^2)$, respectively.
    
    \noindent\textbf{Euclidean distance} is used to the deterministic counterpart to the probabilistic distance.
    The distance between two probability distributions are computed via Monte-Carlo estimation:
    \begin{equation}
        ED(p, q) = \frac{1}{K^2} \sum_{k} \sum_{k'} d(z_p^{(k)}, z_q^{(k')}).
    \end{equation}

    \noindent\textbf{Kullback-Leibler (KL) divergence}~\cite{kl-divergence}
    measures the relative entropy of given two probability distributions as follows:
    \begin{equation}
        \begin{aligned}
            KL(p, q) &= \int \log{\frac{p}{q}} dp   \\
                    &= \frac{1}{2}[\log{\frac{\sigma_q^2}{\sigma_p^2} + \frac{\sigma_p^2}{\sigma_q^2} + \frac{ED(p, q)}{\sigma_q^2}}].
        \end{aligned}
    \end{equation}
    To enforce a distribution-wise symmetric measure, we employ \textbf{Jensen-Shannon (JS) divergence}~\cite{js-divergence} instead of directly using KL-divergence:
    \begin{equation}
        JS(p,q) = \frac{1}{2}[KL(p, q) + KL(q, p)].
    \end{equation}
    
    \noindent\textbf{Wasserstein distance} measures the probability distance of two distributions on a given metric space $M$.
    The 2-Wasserstein distance between two Gaussian distributions is defined as:
    \begin{equation}
        W(p,q)^2 = ED(p,q) + \sigma_p - \sigma_q^2.
    \end{equation}
    
    \begin{table}[]
    \centering
    \small
    \begin{tabular}{
    >{\raggedright}m{0.3\linewidth}
    >{\centering}m{0.1\linewidth}>{\centering}m{0.1\linewidth}
    >{\centering}m{0.1\linewidth}>{\centering}m{0.1\linewidth}
    }
    \hlinewd{0.8pt}
    \multicolumn{5}{l}{\textbf{Number of clips} \quad ($\beta = 10^{-4}$, $K = 10$, $B = 40$)}                              \tabularnewline    
    Parameter $N$                &          1            &         2             &          3            &          4            \tabularnewline
    \hline
    Acc. (\%)                &          78.3            &         81.8             &          82.2            &          \textbf{82.4}            \tabularnewline    \hlinewd{0.8pt}
    \multicolumn{5}{l}{\textbf{Batch size}  \quad ($\beta = 10^{-4}$, $K = 10$, $N = 2$)} \tabularnewline
    Parameter $B$                &          24            &         48             &          72            &          96            \tabularnewline
    \hline
    Acc. (\%)                &          79.9            &         82.1             &          84.5            &          \textbf{86.1}            \tabularnewline   
    \hlinewd{0.8pt}
    \end{tabular}
    \vspace{-7pt}
    \caption{Ablation studies for the number of clips $M$ and batch size $B$. We report the action recognition performance evaluated on UCF101~\cite{ucf101} dataset.}\label{tab:ablation-supp}
    \end{table}
    
    \begin{table}[]
    \centering
    \small
    \begin{tabular}{@{\extracolsep{2pt}}cccccc@{}}
    \hlinewd{0.8pt}
    \multirow{2}{*}{Mining}  &  Recognition   &         \multicolumn{4}{c}{Retrieval}             \\
    \cline{2-2} \cline{3-6}
                            &          Acc. (\%)    &         R@1   &         R@5   &         R@10   &         R@20             \\
    \hline
    \xmark                &          84.2            &         61.2     &         75.5     &         84.0     &         88.7     \tabularnewline 
    \cmark                &          \textbf{86.1}            &         \textbf{64.7}        &         \textbf{78.5}        &         \textbf{87.9}        &         \textbf{91.1}     \tabularnewline
    \hlinewd{0.8pt}
    \end{tabular}
    \vspace{-7pt}
    \caption{Ablation study for positive and negative mining. We evaluate the action recognition and video retrieval performance on the UCF101~\cite{ucf101} dataset.}\label{tab:mining}
    \end{table}

    \subsection{Comparison on Toy Experiment}
    
    Now we compare the mining capacity through the toy experiment on 11 subclasses from UCF101~\cite{ucf101} dataset\footnote{The 11 classes (``ApplyEyeMakeup", ``Archery", ``BabyCrawling", ``BalanceBeam", ``BandMarching", ``BaseballPitch", ``Basketball", ``BenchPress", ``Biking", ``GolfSwing", and ``SkyDiving") are used for this experiment.}.
    We construct a batch by sampling 8 videos per class such that the batch size is 88 and the number of hard positive pairs (having the same class) is 308.
    Different from the main text, we set threshold distance $\tau$ as the average value of self-distances:
    \begin{equation}
        \tau = \frac{1}{B} \sum_{i=1}^{B} \text{dist}(\mathcal{V}_i, \mathcal{V}_i),
    \end{equation}
    where $B$ is batch size and $\text{dist}(\mathcal{V}_i, \mathcal{V}_i)$ represents the average distance between embedding pairs sampled from the video distribution $p(z|\mathcal{V}_i)$.
    We compute the precision and recall at every 20 epochs for the constructed positive pairs according to the type of distance to estimate the mining capacity, as shown in \figref{fig:precision-recall}.
    At the beginning of training, high recall is shown on the right side of \figref{fig:precision-recall} regardless of the distance due to a large number of positive candidates, while the precision is significantly low, as shown on the left side of \figref{fig:precision-recall}.
    The comparison between Euclidean distance and the probabilistic distances implies that Euclidean distance requires sufficient training steps to eliminate false positives, showing high recall and low precision compared to the probabilistic distance.
    The results on the probabilistic distance show that Bhattacharyya distance achieves constructing reliable positive pairs with a small number of epochs.
    
    \begin{table}[]
    \centering
    \small
    \begin{tabular}{
    >{\raggedright}m{0.3\linewidth}
    >{\centering}m{0.1\linewidth}>{\centering}m{0.1\linewidth}>{\centering}m{0.1\linewidth}>{\centering}m{0.1\linewidth}
    }
    \hlinewd{0.8pt}
    Similarity Metric                        &         R@1   &         R@5   &         R@10   &         R@20             \tabularnewline
    \hline
    Cosine Similarity                &         63.1     &         77.0     &         86.3     &         89.8     \tabularnewline 
    Match Probability               &         \textbf{64.7}        &         \textbf{78.5}        &         \textbf{87.9}        &         \textbf{91.1}     \tabularnewline
    \hlinewd{0.8pt}
    \end{tabular}
    \caption{Ablation study for the similarity metric on video retrieval. We evaluate video retrieval performance on UCF101~\cite{ucf101} dataset.}\label{tab:metric}
    \end{table}

    \section{Implementation Details}\label{sec:3}
\subsection{Two-stage Training}
        In pretraining, our ProViCo trains the model in a two-stage procedure for stable training following \cite{co-crl, mfo}.
        At the first 30 epochs, the model is trained without hard positive mining such that the positive pairs are defined as
        \begin{equation}
            \mathcal{P} = \lbrace (\mathcal{V}_i, \mathcal{V}_j) \;|\; i = j\rbrace.
        \end{equation}
        After initial training, the positive and negative pairs are selected based on the probabilistic distance as described in the main text.
        The two-stage training enables the model to obtain more substantial initial parameters than randomly initialized models and construct the confident training pairs.
        % With the two-stage training, the model obtains stronger initialized parameters than randomly initialized models and constructs the confident training pairs.
    
\subsection{Finetuning and Inference}
\noindent\textbf{Finetuning.}
    After self-supervised pretraining, we finetune the model on UCF101~\cite{ucf101} or HMDB51~\cite{hmdb51} datasets for action recognition.
    As same with the pretraining, we randomly sample two 16-frame clips with the temporal stride of 1 from each video and all frames are fixed to a size of $112 \times 112$ by random cropping.
    The backbone network with an additional fully-connected (FC) layer are trained for 200 epochs with a mini-batch size of 96 and the learning rate of $0.02$.
    To predict the action class of a video $\mathcal{V}$, the FC layer takes all embeddings $\lbrace{z^{(k)}}\rbrace_{k=1}^{K}$ sampled from $p(z|\mathcal{V})$ as inputs and outputs the sample-wise class probabilities $\lbrace{y^{(k)}}\rbrace_{k=1}^{K}$ followed by a softmax function.
    We apply cross-entropy loss between the averaged probability score $\frac{1}{K} \sum_k y^{(k)}$ and the groundtruth to learn parameters.
  
    \begin{figure*}[t]
        \centering
            \begin{subfigure}{0.18\linewidth}{\includegraphics[width=1.0\linewidth]{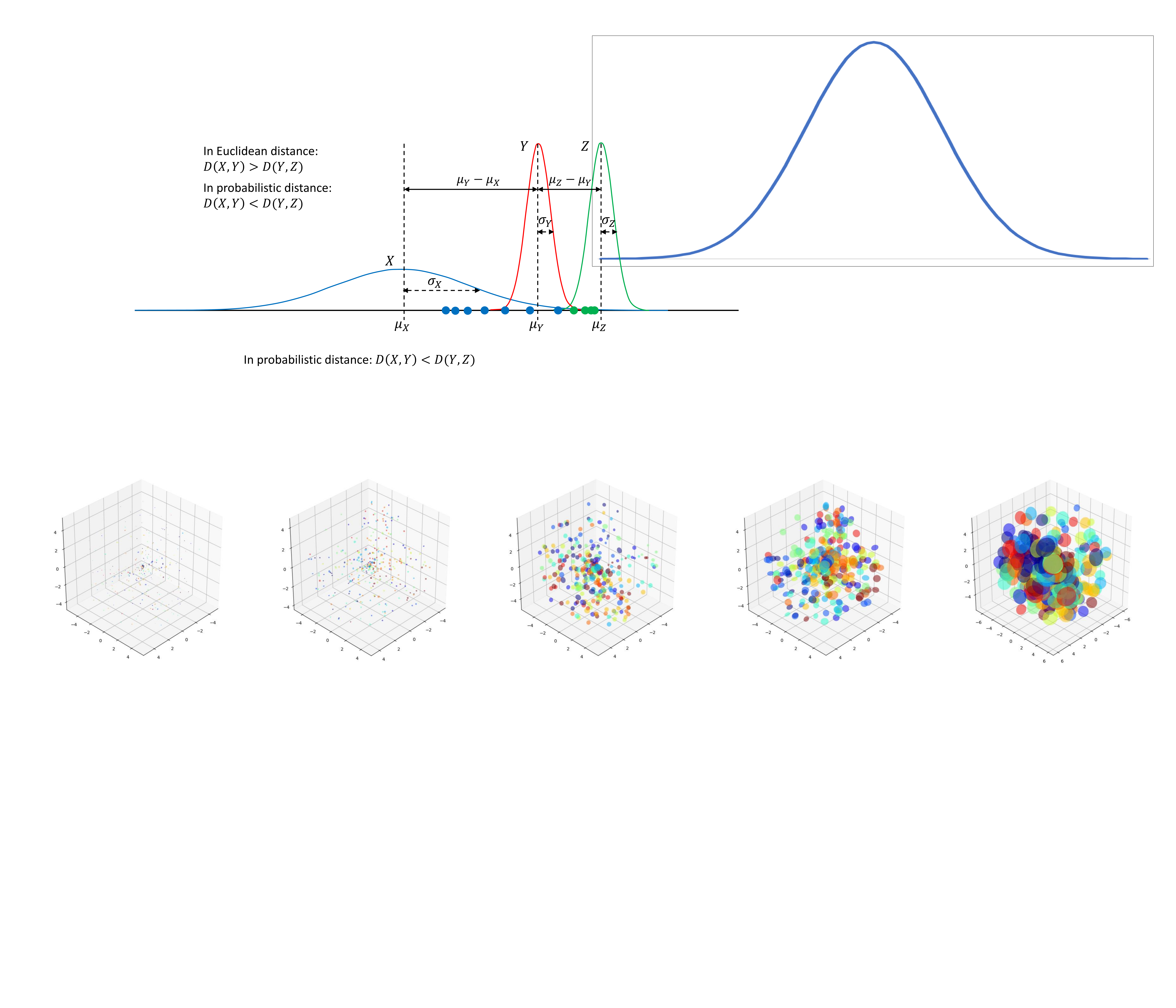}}
            \caption{$\beta=10^{-6}$}\label{fig:betaa}
            \end{subfigure}
            \hspace{3pt}
            \begin{subfigure}{0.18\linewidth}{\includegraphics[width=1.0\linewidth]{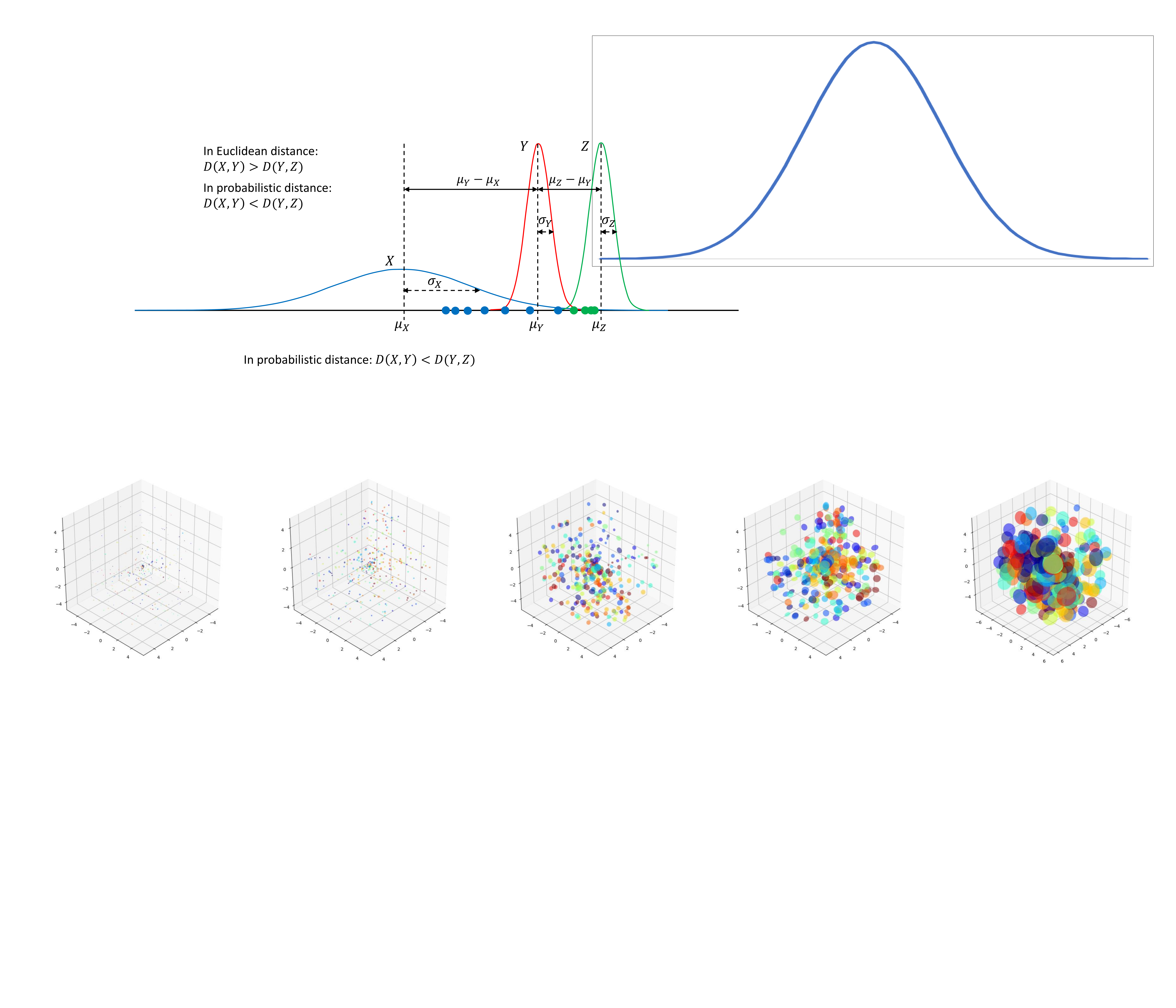}}
            \caption{$\beta=10^{-5}$}\label{fig:betab}
            \end{subfigure}
            \hspace{3pt}
            \begin{subfigure}{0.18\linewidth}{\includegraphics[width=1.0\linewidth]{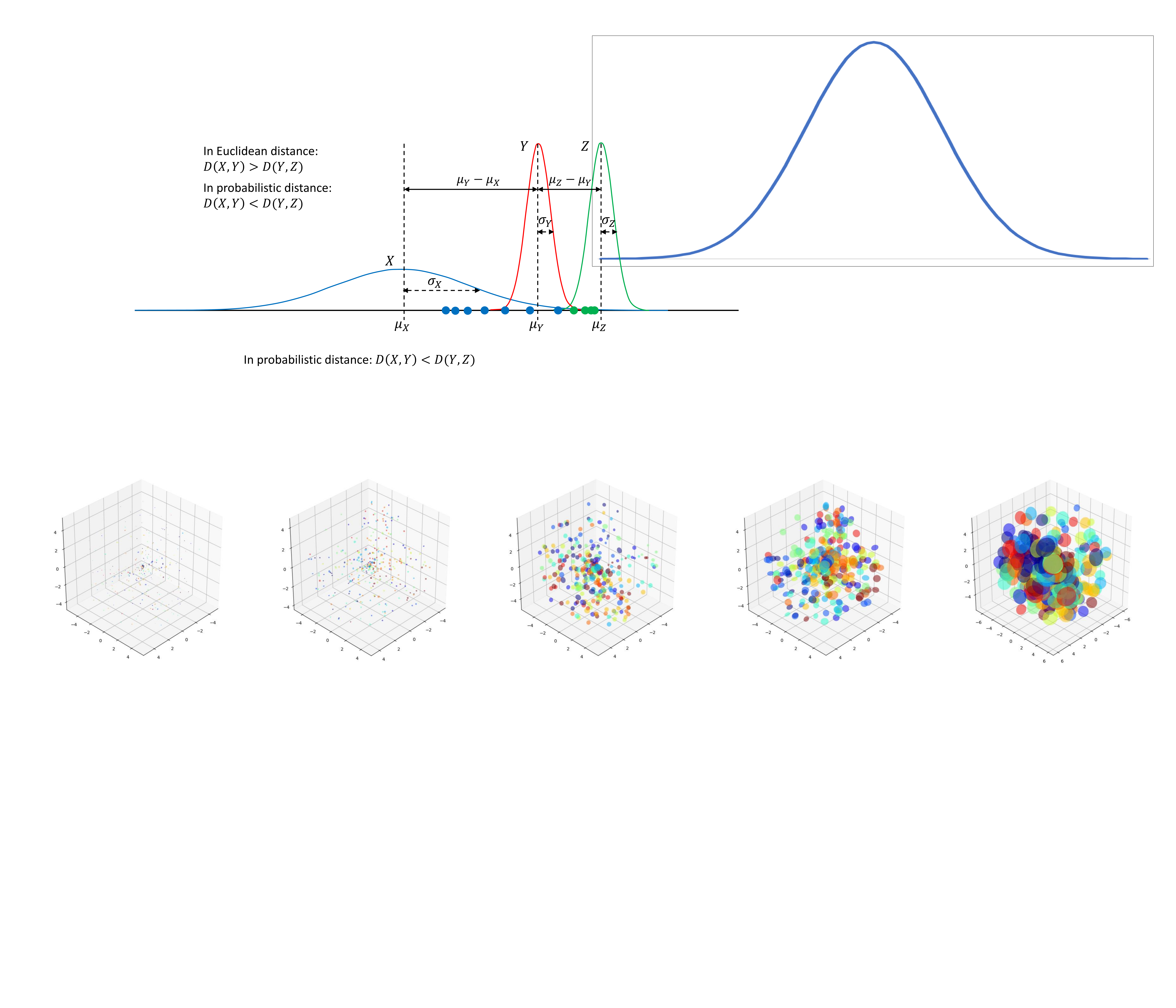}}
            \caption{$\beta=10^{-4}$}\label{fig:betac}
            \end{subfigure}
            \hspace{3pt}
            \begin{subfigure}{0.18\linewidth}{\includegraphics[width=1.0\linewidth]{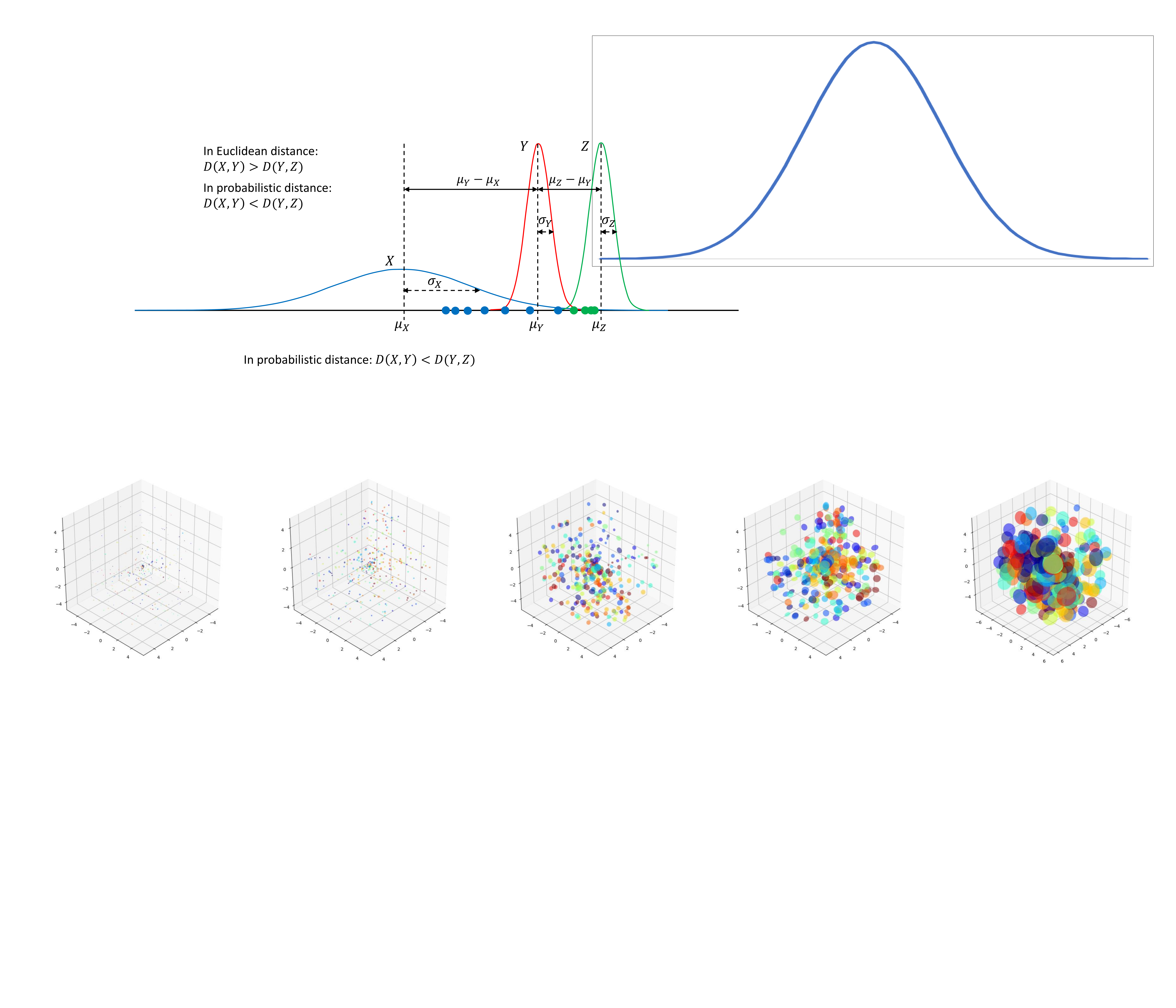}}
            \caption{$\beta=10^{-3}$}\label{fig:betad}
            \end{subfigure}
            \hspace{3pt}
            \begin{subfigure}{0.18\linewidth}{\includegraphics[width=1.0\linewidth]{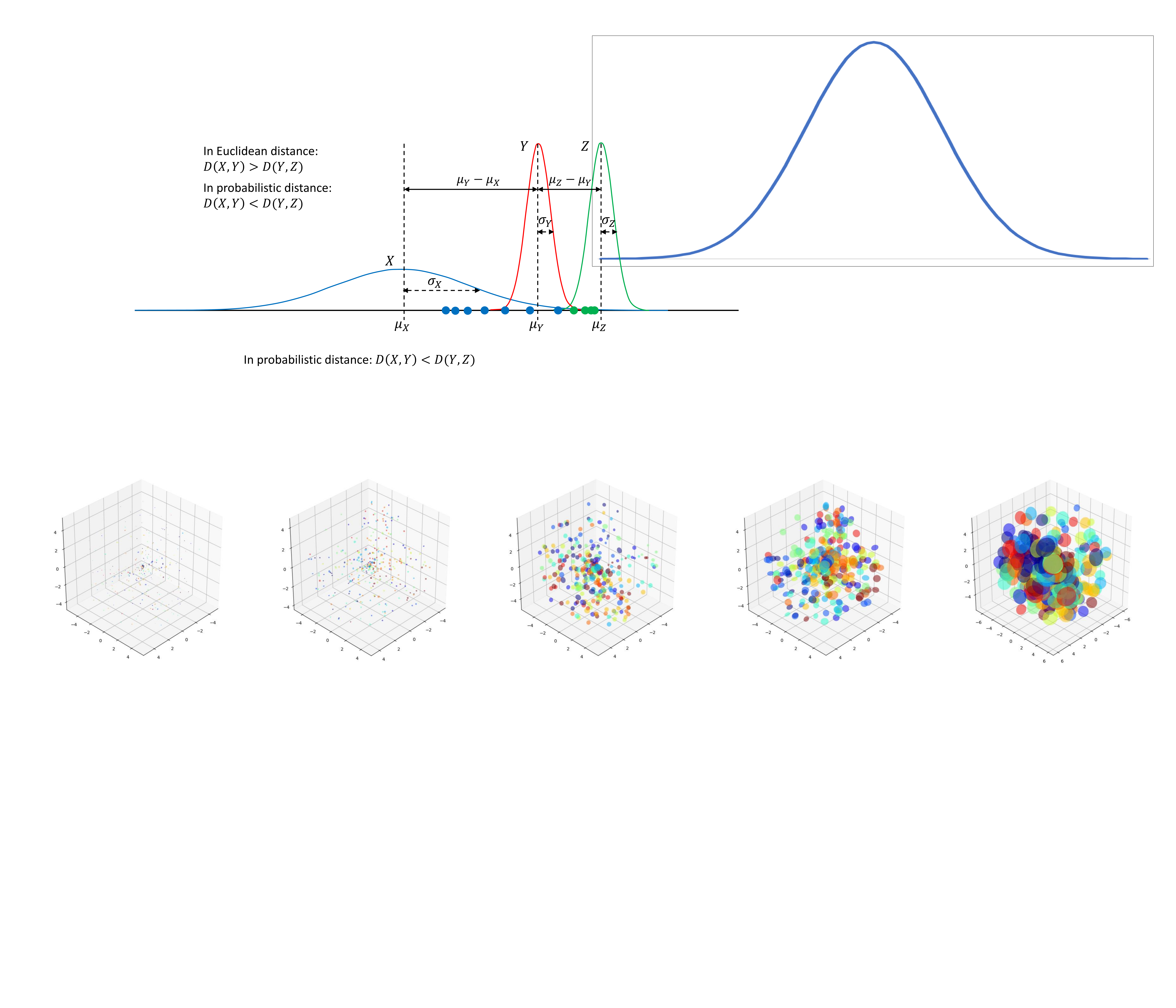}}
            \caption{$\beta=10^{-2}$}\label{fig:betae}
            \end{subfigure}
            
        \caption{
        \textbf{Impact of the KL-divergence hyperparameter $\beta$.} We visualize 3-dimensional embeddings learned with the different values of $\beta$ on 11 subclasses of UCF101~\cite{ucf101} dataset. Each class is viewed in color.}\label{fig:beta}
    \end{figure*}

  \begin{figure*}[!ht]
    \begin{center}
       \includegraphics[width=1\linewidth]{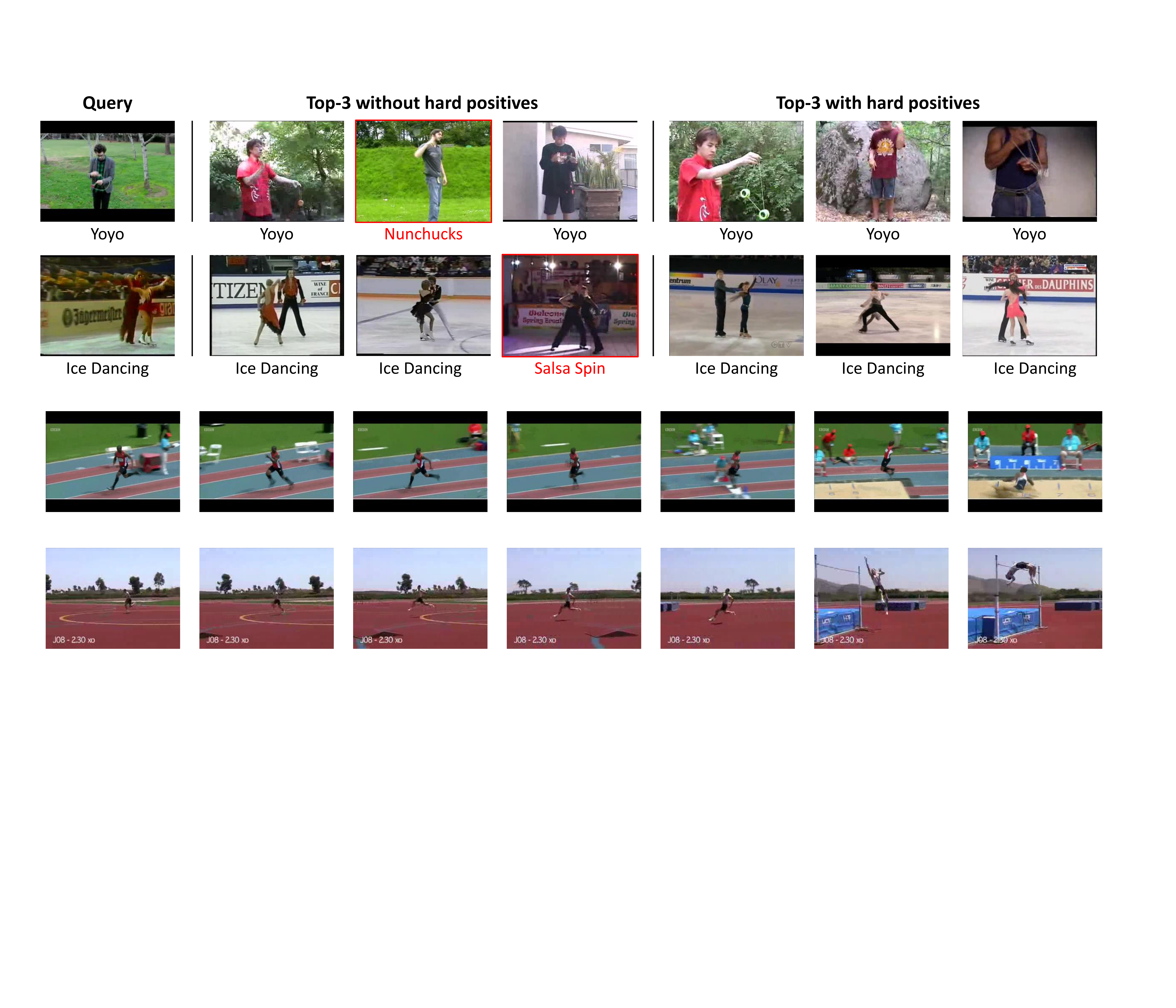} \hfill \\
    \end{center}
    \vspace{-18pt}
      \caption{\textbf{Qualitative results for video retrieval.} (Left) Input query videos, (Middle) top-3 nearest-neighbors retrieved by the model trained without hard positives, and (Right) top-3 nearest-neighbors retrieved by the model trained with hard positives. The red box indicates the wrong retrieval results.}
    \label{fig:retrieval}
    \end{figure*}
    
\noindent\textbf{Inference.}
    For action recognition, we uniformly sample 10 clips for each video and apply center cropping for fixed size of $112 \times 112$, following \cite{mfo, pp}.
    Specifically, the consecutive two clips are used to estimate the video distribution, such that five video distributions are estimated from 10 clips.
    We sample $K$ embeddings on each distribution and predict $5K$ class probabilities.
    Therefore, the final prediction of each video is the averaged probability of $5K$ embeddings.
    For video retrieval task, we basically follow the experimental protocol in~\cite{mfo, mcn} and use the pretrained model without any additional training.
    Each video in the test split is taken as a query and top-k nearest-neighbors are retrieved from the training set.
    To this end, the match probability~\cite{hib, pcme} is employed to measure the similarity, and the retrieval performance is evaluated by top-k R@k.
    
\section{Additional Results}\label{sec:4}

    In this section, we provide additional experimental results with ProViCo.
    The results include ablation studies, 3D visualization, and qualitative results for video retrieval.

\subsection{Ablation Study}
    We provide additional ablation studies for the number of clips $N$ used during training, batch size $B$, effectiveness of positive and negative mining, and the similarity metrics.
    
\noindent\textbf{Number of clips and batch size.}
    We report the action recognition performance evaluated on UCF101~\cite{ucf101} dataset corresponding to the number of input clips $N$ used during training in the first block of \tabref{tab:ablation-supp}.
    With the limited GPU memory, we set batch size to 40 for all results to eliminate the effect from batch size.
    Since more sophisticated and complicated distribution for the whole video can be estimated by combining a larger number of clip distributions, the performance is improved as $N$ increases.
    However, the results that two clips are sufficient to represent the whole video distribution, showing competitive performance to the larger number of clips.
    In addition, we observe that batch size has a more impact on the performance than the number of clips as shown in the second block of \tabref{tab:ablation-supp}.
    To consider the trade-off in terms of the computational capacity and the space complexity, we use two clips ($N=2$) for each video in the main experiments.
    This also provides fair comparisons with previous methods~\cite{cvrl, tec} under general experimental settings.

\noindent\textbf{Positive and negative mining.}
    We ablate positive and negative mining to verify the effectiveness of the hard positive pairs by evaluating the action recognition and video retrieval performances on UCF101~\cite{ucf101} dataset.
    As shown in \tabref{tab:mining}, constructing the positive and negative pairs based on the probabilistic distance obviously improves the performance of downstream tasks, especially on video retrieval.
    
    \begin{figure*}[!ht]
        \centering
            \begin{subfigure}{0.18\linewidth}{\includegraphics[width=1.0\linewidth]{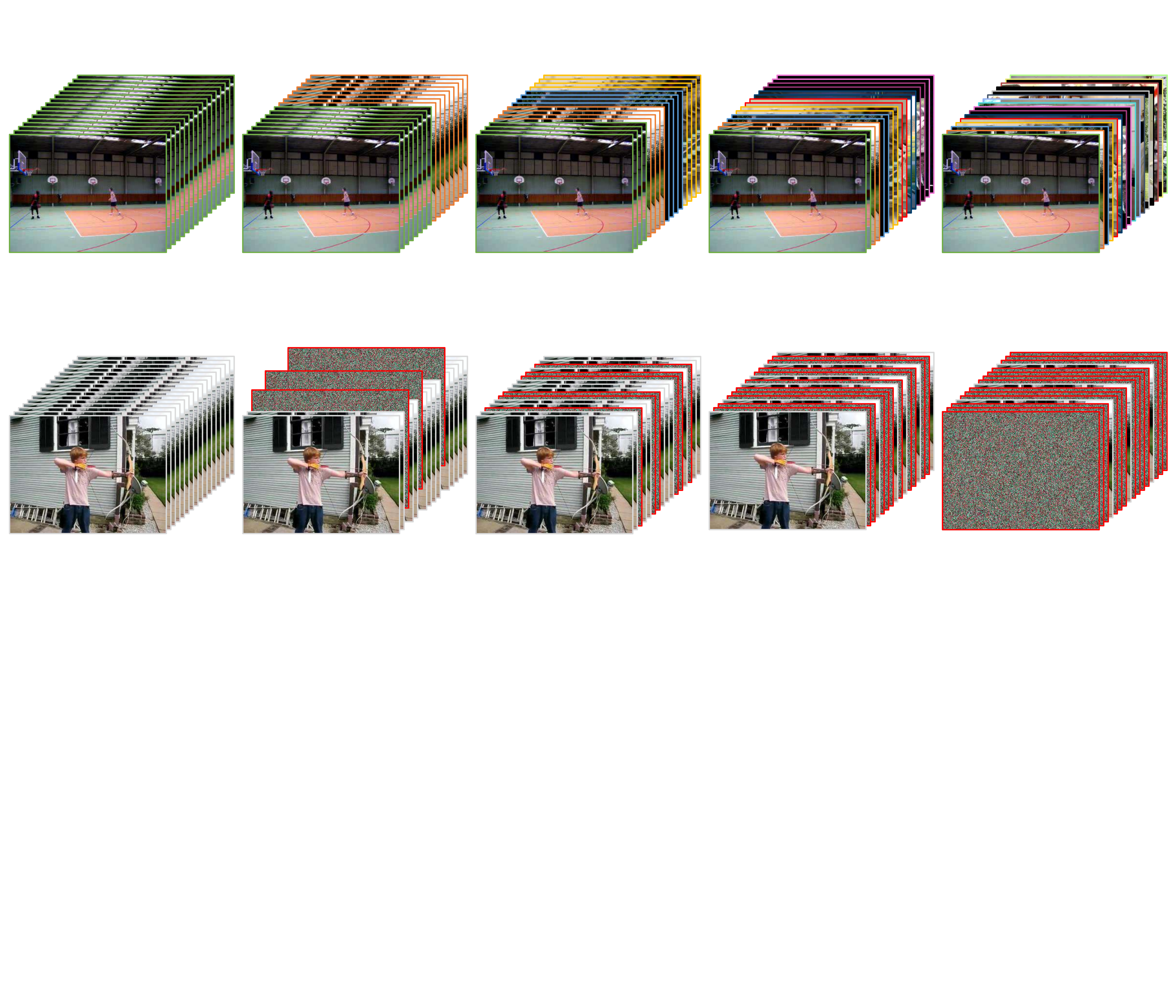}}\hfill
            \caption{Level 1}\label{fig:levela}
            \end{subfigure}
            \hspace{3pt}
            \begin{subfigure}{0.18\linewidth}{\includegraphics[width=1.0\linewidth]{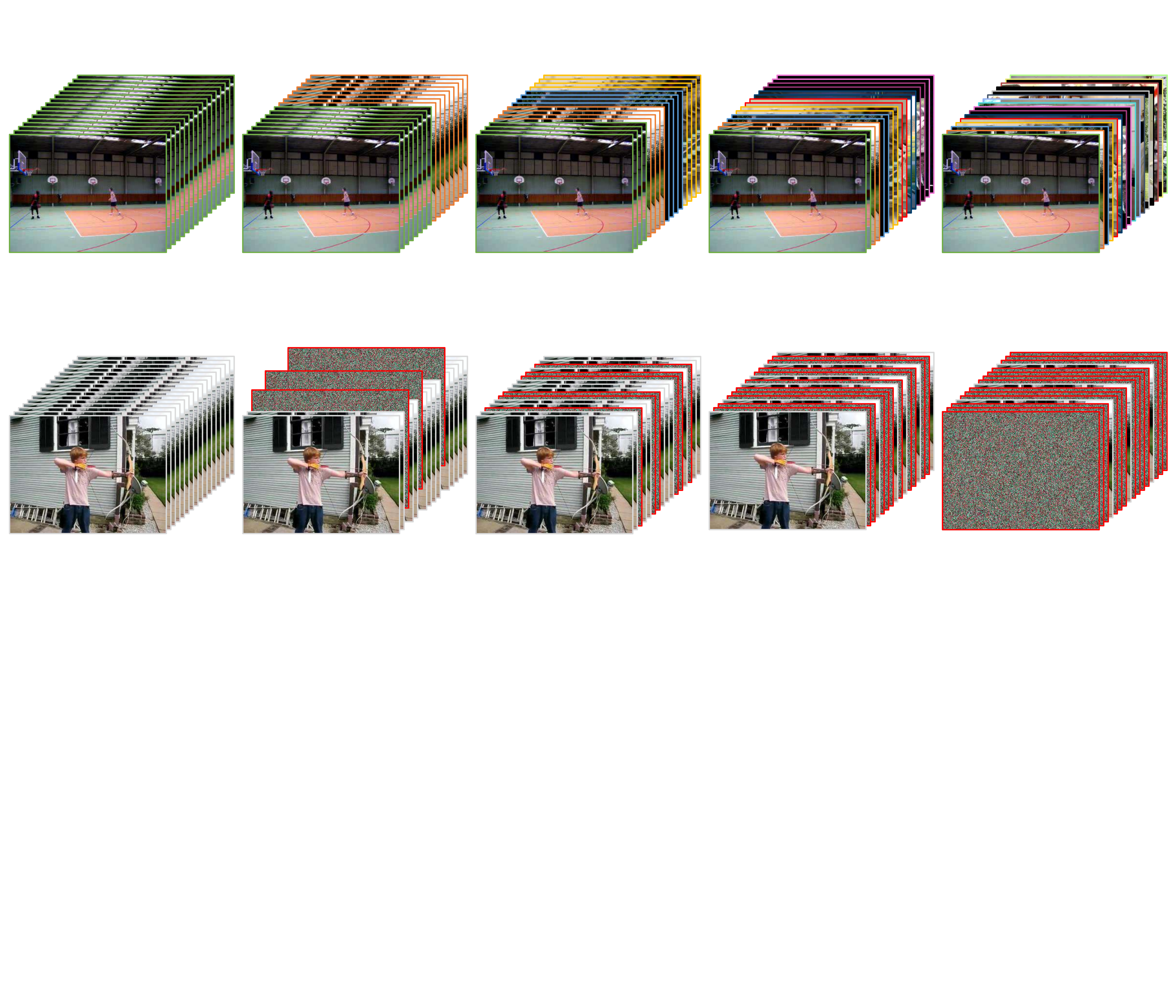}}\hfill
            \caption{Level 2}\label{fig:levelb}
            \end{subfigure}
            \hspace{3pt}
            \begin{subfigure}{0.18\linewidth}{\includegraphics[width=1.0\linewidth]{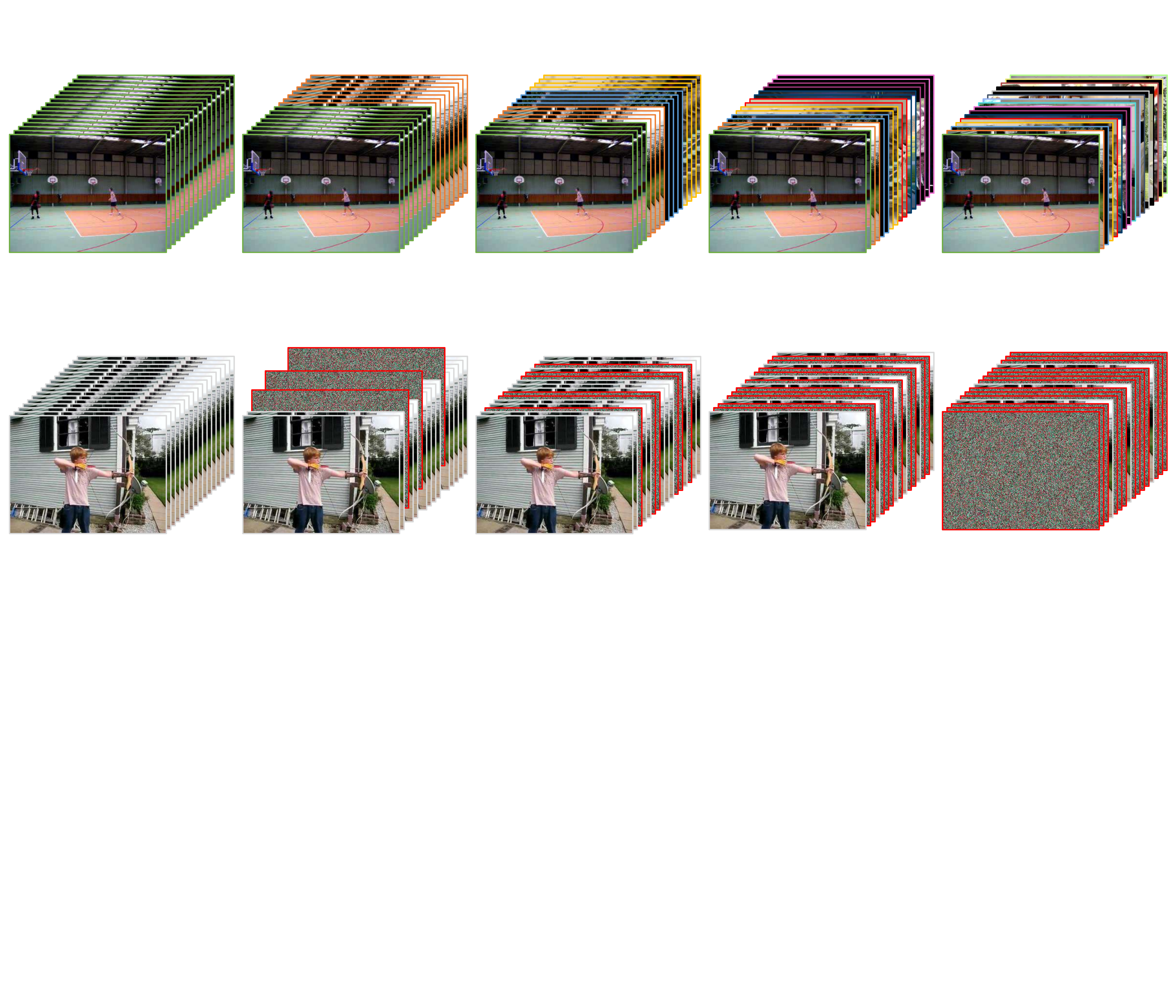}}\hfill
            \caption{Level 3}\label{fig:levelc}
            \end{subfigure}
            \hspace{3pt}
            \begin{subfigure}{0.18\linewidth}{\includegraphics[width=1.0\linewidth]{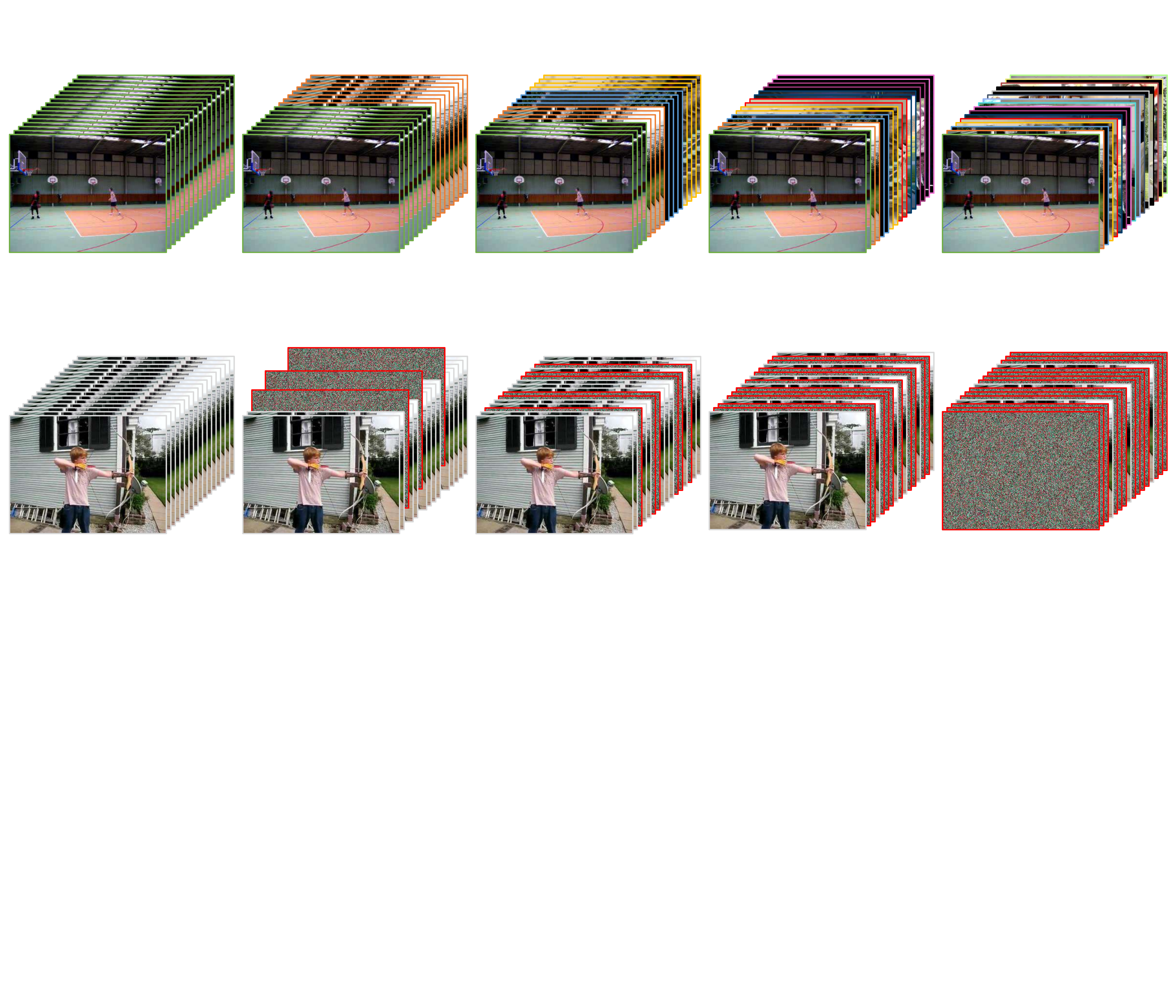}}\hfill
            \caption{Level 4}\label{fig:leveld}
            \end{subfigure}
            \hspace{3pt}
            \begin{subfigure}{0.18\linewidth}{\includegraphics[width=1.0\linewidth]{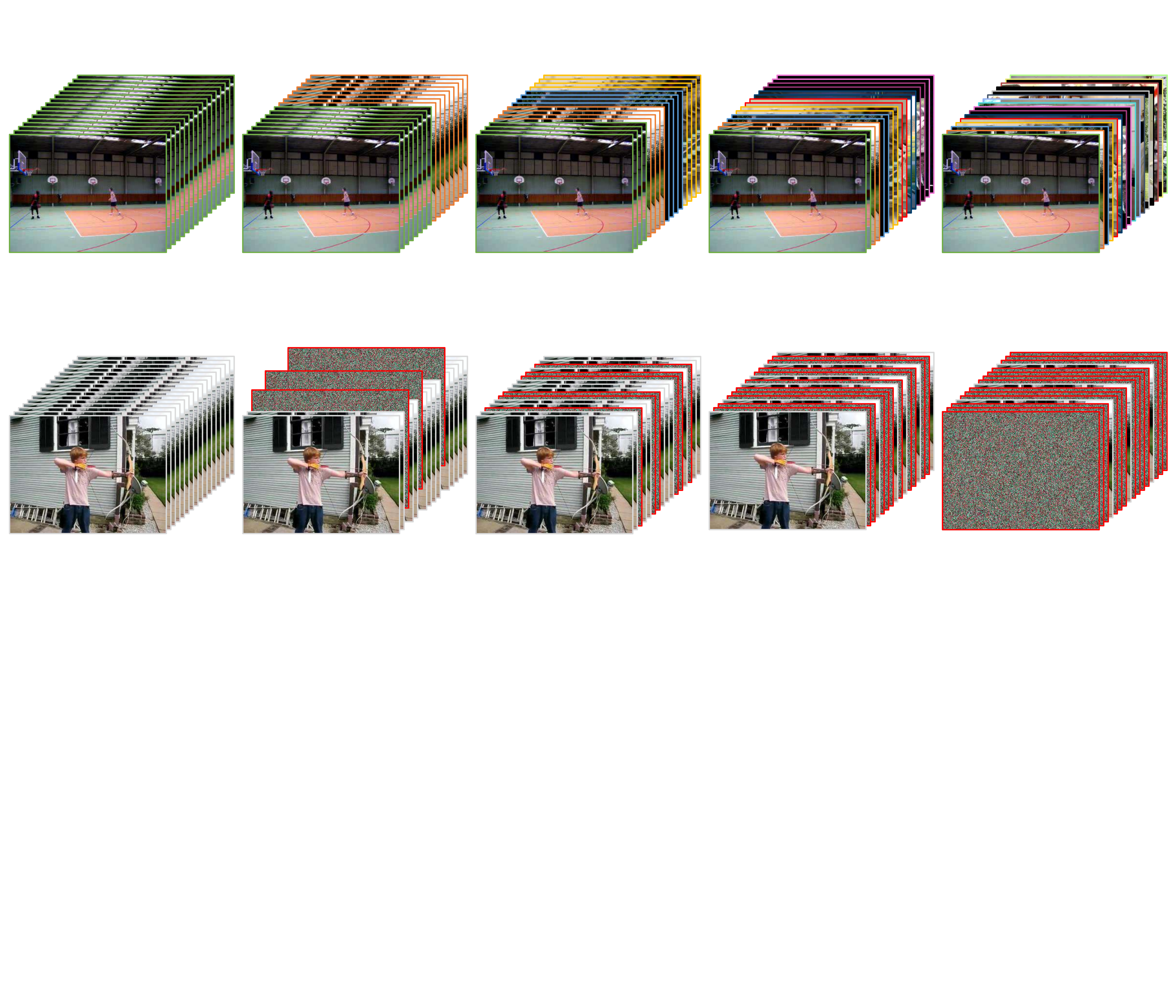}}\hfill
            \caption{Level 5}\label{fig:levele}
            \end{subfigure}

        \vspace{-10pt}
        \caption{\textbf{Generation of mixed (uncertain) clips.} The uncertainty divided into five levels according to the number of videos used to generate mixed clips.}\label{fig:mixed}
    \end{figure*}
    
    \begin{figure*}[!ht]
        \centering
            \begin{subfigure}{0.18\linewidth}{\includegraphics[width=1.0\linewidth]{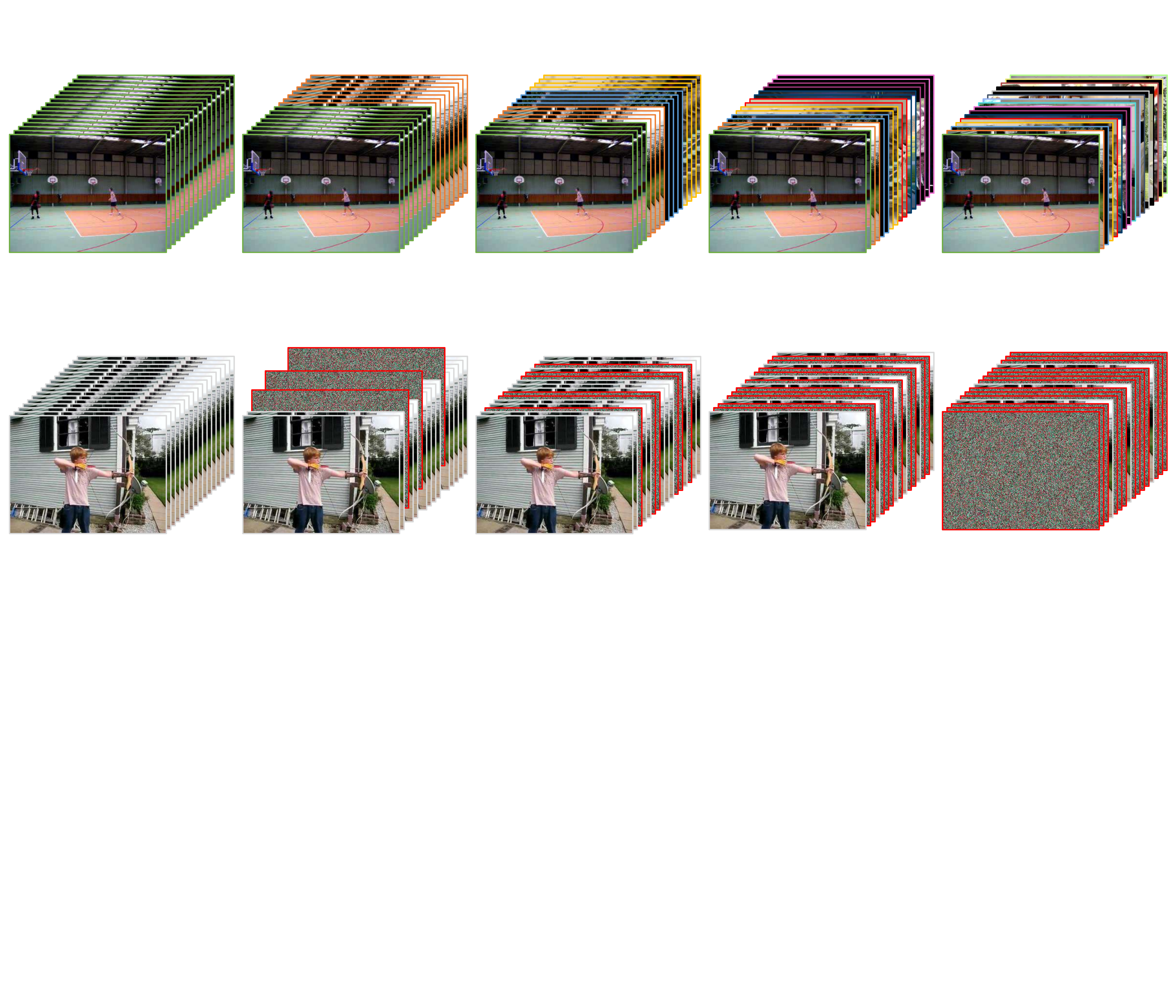}}\hfill
            \caption{$\rho=0$}\label{backa}
            \end{subfigure}
            \hspace{3pt}
            \begin{subfigure}{0.18\linewidth}{\includegraphics[width=1.0\linewidth]{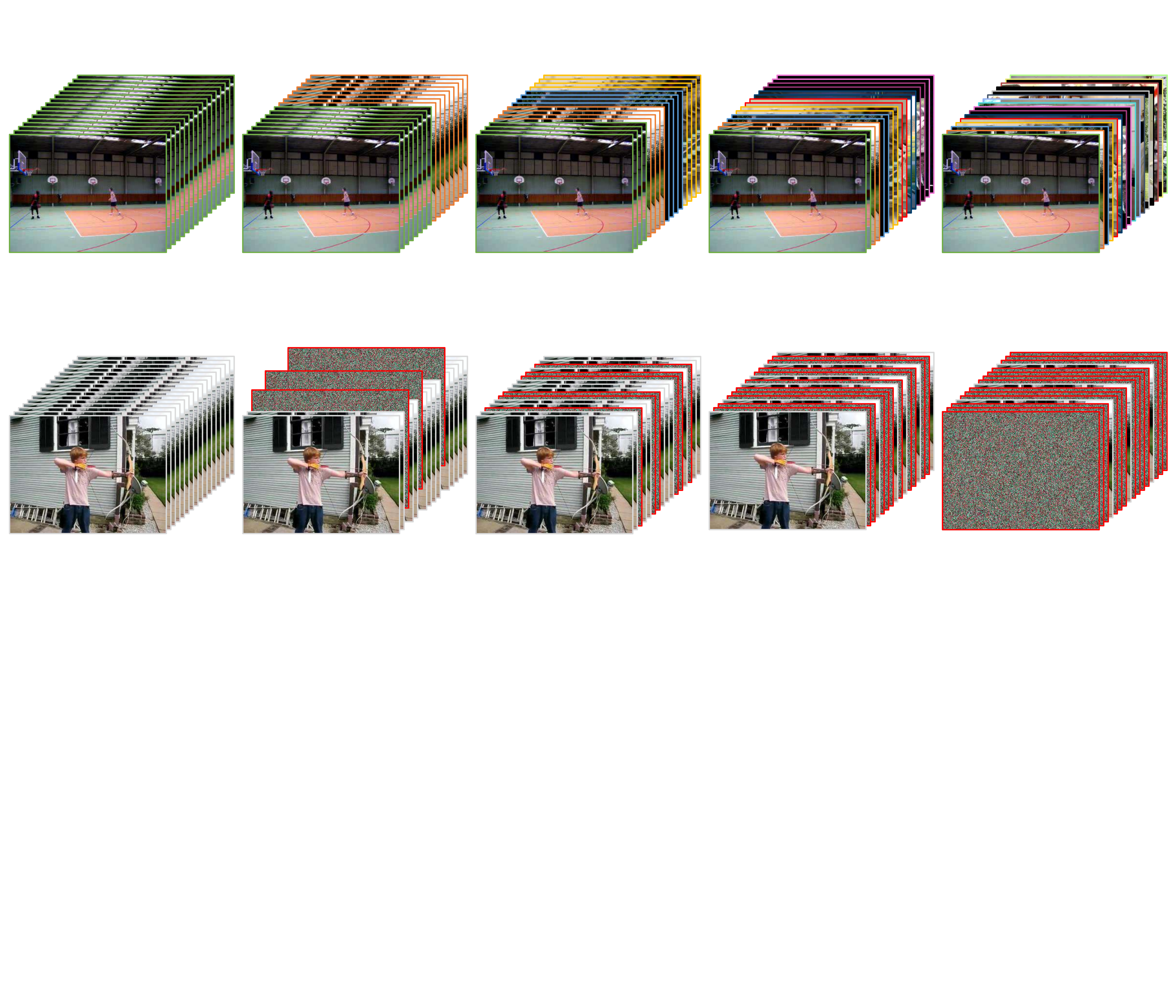}}\hfill
            \caption{$\rho=0.2$}\label{backb}
            \end{subfigure}
            \hspace{3pt}
            \begin{subfigure}{0.18\linewidth}{\includegraphics[width=1.0\linewidth]{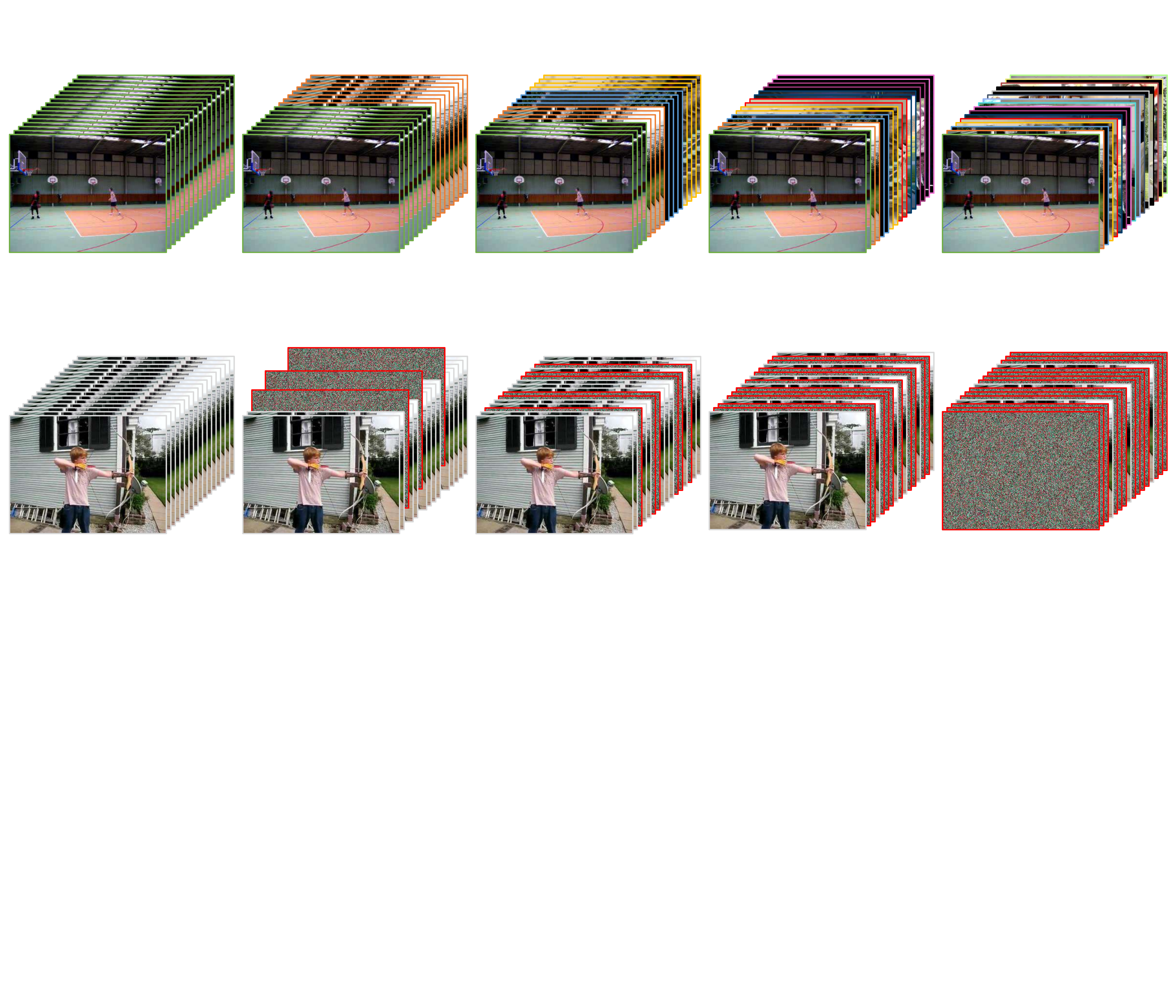}}\hfill
            \caption{$\rho=0.4$}\label{backc}
            \end{subfigure}
            \hspace{3pt}
            \begin{subfigure}{0.18\linewidth}{\includegraphics[width=1.0\linewidth]{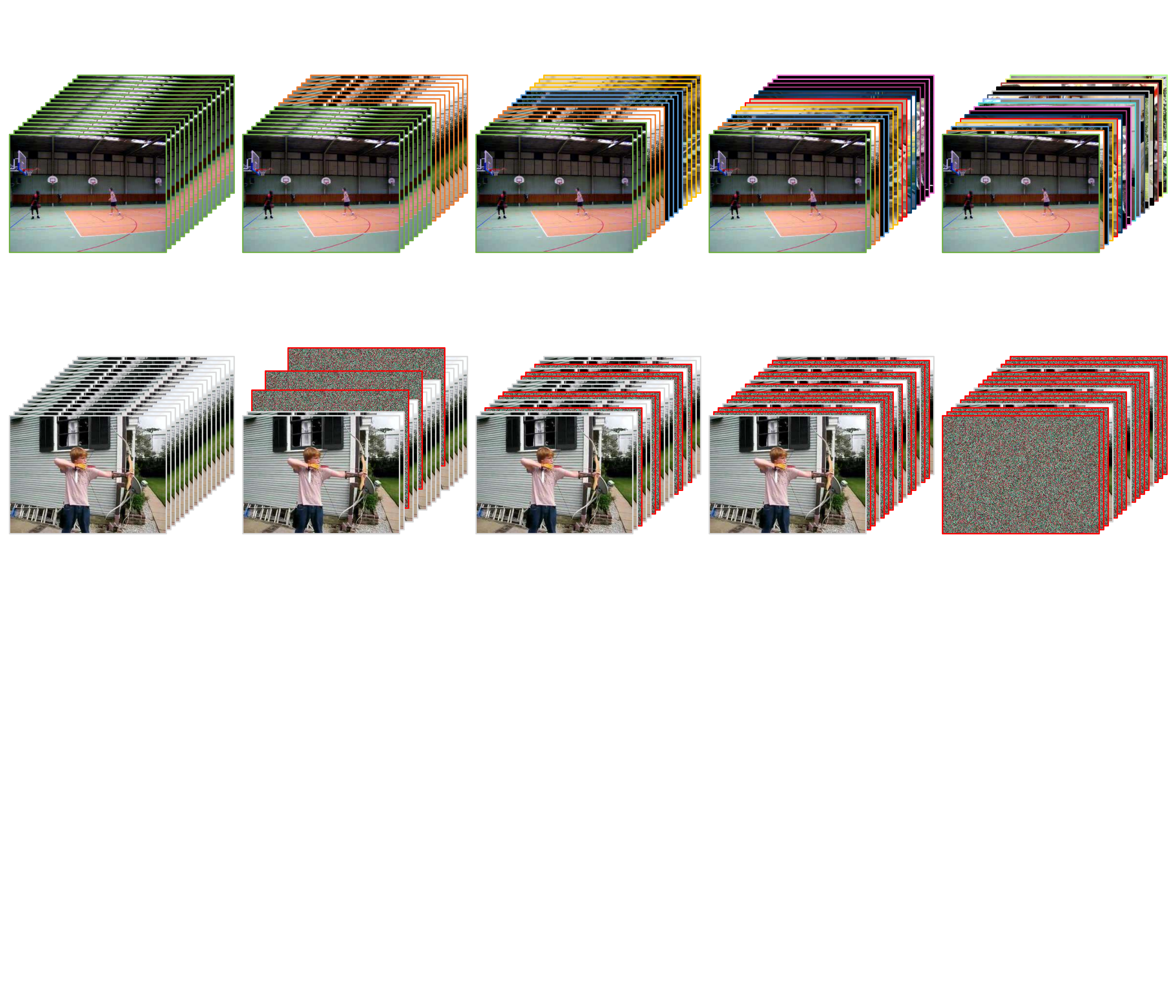}}\hfill
            \caption{$\rho=0.6$}\label{backd}
            \end{subfigure}
            \hspace{3pt}
            \begin{subfigure}{0.18\linewidth}{\includegraphics[width=1.0\linewidth]{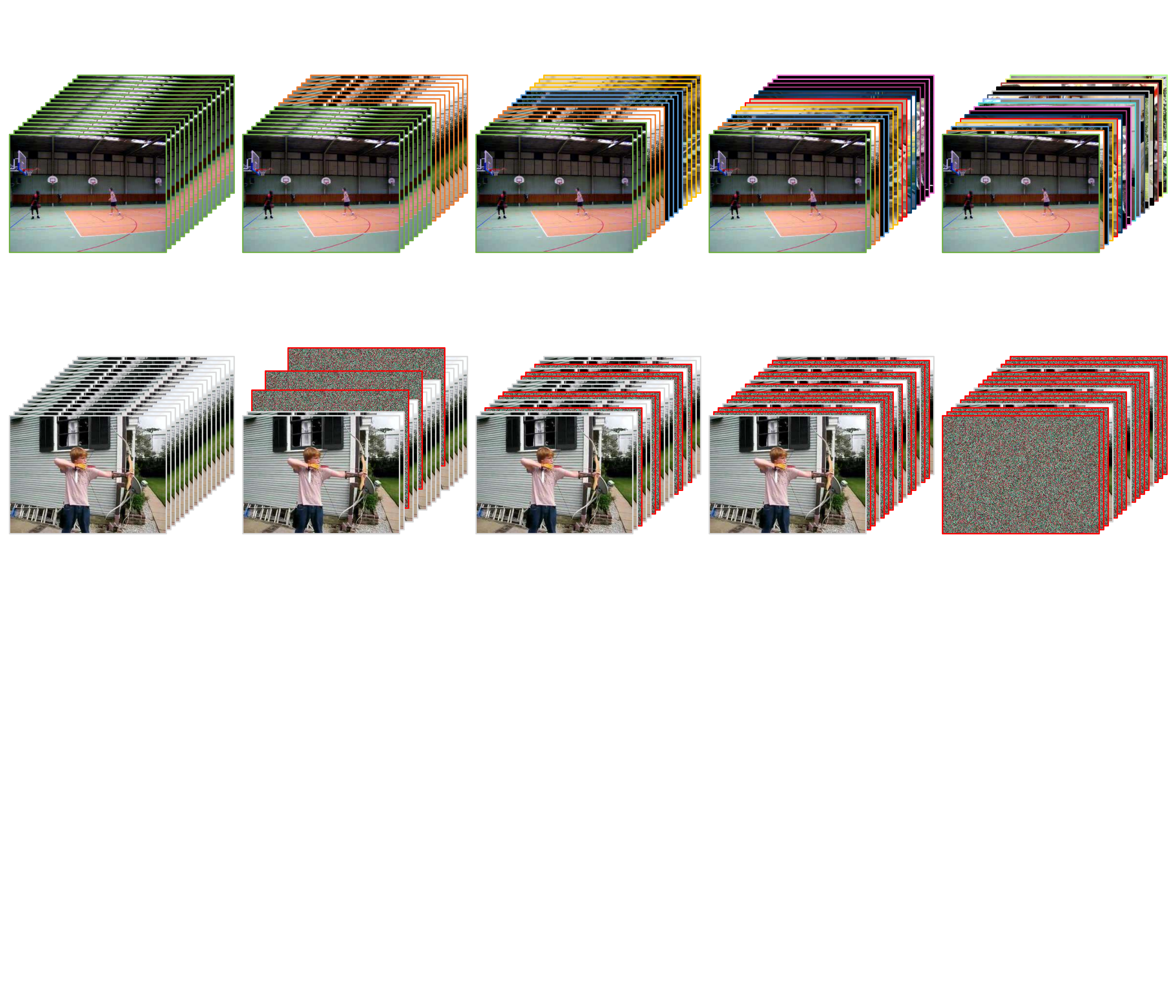}}\hfill
            \caption{$\rho=0.8$}\label{backe}
            \end{subfigure}
            
        \vspace{-10pt}
        \caption{\textbf{Generation of masked (background) clips.} The frames in the original clip are replaced by noise frames according to the removal ratio $\rho$.}\label{fig:back}
    \end{figure*}
    
    \begin{figure}[t]
    \begin{center}
    %\fbox{\rule{0pt}{2in} \rule{0.9\linewidth}{0pt}}
       \includegraphics[width=1\linewidth]{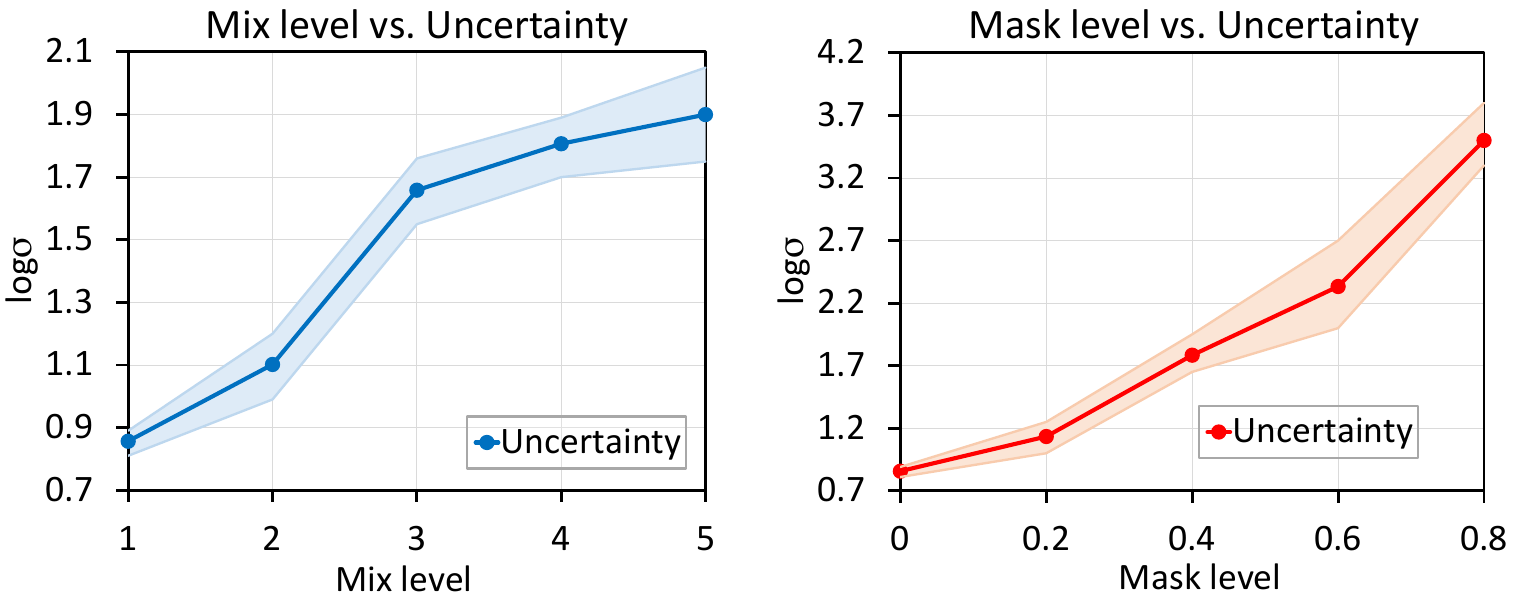}
    \end{center}
    \vspace{-15pt}
      \caption{\textbf{(Left) Mix level versus uncertainty} and \textbf{(Right) Mask level versus uncertainty.} We measure the uncertainty of videos on three test splits of UCF101~\cite{ucf101} dataset according to each corrupted level.}
    %   \vspace{-7pt}
    \label{fig:uncertainty-level}
    \end{figure}
    
\noindent\textbf{Similarity metric on video retrieval.}
    We mainly use the match probability~\cite{hib, pcme} as a similarity metric to retrieve videos.
    To verify the performance on variants of the similarity metric, we evaluate the video retrieval performance in \tabref{tab:metric}.
    For the cosine similarity, we use the mean vector of each video to measure the similarity without embedding sampling.
    The results show that the performance using the match probability is better than the cosine similarity.
    On the other hand, in terms of space complexity, the cosine similarity and match probability require $\mathcal{O}(N)$ and $\mathcal{O}(K^2N)$ spaces, respectively. 
    Therefore, we can alternatively use the match probability to obtain more accurate retrieval performance and the cosine similarity for the fast inference.

    \section{Extension with Uncertainty}\label{sec:5}

    \subsection{KL Regularization with 3D Visualization}
    To visually observe the learned representations and the impact of the KL-divergence hyperparameter $\beta$, we conduct an additional toy experiment on 11 subclasses used in \secref{sec:mining}.
    Specifically, we slightly transform the architecture of ProViCo to learn 3-dimensional embeddings.
    We use the same architecture as the main experiments, but add two additional projection layers that take $g_\mu (v_{c_n})$ and $g_\sigma (v_{c_n})$ as an input respectively to obtain 3-dimensional embeddings.
    Note that we use all the videos from three train splits for 11 classes in this experiment.
    In \figref{fig:beta}, we visualize learned embeddings on 3D space according to $\beta$.
    As analyzed in Sec. 4.5 of the main text, embeddings becomes like points at the small value ($\beta = 10^{-6}$).
    On the contrary to this, an increase in $\beta$ leads to an increase in the variance of embeddings, such that embeddings approach to the unit Gaussian at the large value ($\beta = 10^{-2}$).

\subsection{Qualitative Results}
    In \figref{fig:retrieval}, we visualize video retrieval results obtained by top-3 nearest-neighbors on the test split 1 of UCF101~\cite{ucf101} dataset.
    The results show that the model trained with hard positives more robust to the semantic instance discrimination than the model trained without any hard positives.

    As mentioned in Sec. 1 of the main text, our probabilistic framework can make useful applications such as estimation of difficulty or chance of failure on test data.
    In this section, we study the uncertainty with corrupted test videos by establishing two factors that increase the uncertainty of the video:
    % \jy{(1) the ambiguous content of the video which contains frames unrelated to the subject of the video and (2) the loss content of the video which includes the meaningless background frames.}
    (1) the ambiguous content of the video, which is unrelated to the subject of the video, and (2) the empty content of the video, which includes the meaningless background frames.
    
    \subsection{Uncertainty on Mixed Clips}
    To make the content of the video uncertain, we generate clips by mixing frames from several videos.
    We depict examples of generated mixed clips in \figref{fig:mixed}.
    We divide the uncertainty into five levels according to the number of video clips used to generate mixed clips.
    For example, \figref{fig:levelc}, representing mixture level 3, composes a 16-frame clip by mixing each of four consecutive frames sampled from four videos.
    We measure the averaged uncertainty of videos from three test splits on UCF101~\cite{ucf101} dataset according to uncertainty levels.
    As shown on the left side in \figref{fig:uncertainty-level}, the uncertainty increases as the number of mixed videos increases.
    Based on this result, the uncertainty predicted by our method can be used to eliminate the ambiguous clip or balance the weights between clips for reliable performance on various downstream tasks.
    
    \subsection{Uncertainty on Masked Clips}
    To consider the case where the video contains background frames, we randomly remove frames of the clip and insert random noise frames to removed positions.
    We divide the background degree according to the removal ratio $\rho$ (\ie, $\rho = 0, 0.2, 0.4, 0.6, 0.8$), as shown in \figref{fig:back}.
    The results on the right side in \figref{fig:uncertainty-level} show the uncertainty exponentially increases as the background ratio increases.
    In practice, \textit{untrimmed raw videos} consist of sparse frames related to the subject of the video and the majority of meaningless background frames.
    Our probabilistic approach with the uncertainty estimation enables us to effectively exploit the untrimmed videos by filtrating the ambiguous or meaningless contents.

\end{document}